%% file: Main.tex
\documentclass[10pt,twocolumn,letterpaper]{article}

\usepackage[pagenumbers]{wacv}              

\usepackage{verbatim}

\usepackage{graphicx}
\usepackage{amsmath}
\usepackage{amssymb}
\usepackage{booktabs}
\usepackage{float}
\usepackage{svg}
\usepackage{xcolor}
\usepackage[normalem]{ulem}
\usepackage{float}
\usepackage{caption}
\usepackage{subcaption}
\usepackage{colortbl}
\useunder{\uline}{\ul}{}

\definecolor{gray}{rgb}{0.9,0.9,0.9}

\usepackage[pagebackref,breaklinks,colorlinks]{hyperref}
\usepackage{orcidlink}

\usepackage[capitalize]{cleveref}
\crefname{section}{Sec.}{Secs.}
\Crefname{section}{Section}{Sections}
\Crefname{table}{Table}{Tables}
\crefname{table}{Tab.}{Tabs.}

\def\wacvPaperID{2266} 
\def\confName{WACV}
\def\confYear{2025}

\begin{document}
\include{Paper}
{\small
\bibliographystyle{ieee_fullname}
\bibliography{egbib}
}
\clearpage
\setcounter{section}{0}
\renewcommand{\thesection}{\Alph{section}}
\include{Suppl}
\end{document}

%% file: Paper.tex
\title{I Dream My Painting: Connecting MLLMs and Diffusion Models via Prompt Generation for Text-Guided Multi-Mask Inpainting}


\author{
Nicola Fanelli\orcidlink{0009-0007-6602-7504} \qquad Gennaro Vessio\orcidlink{0000-0002-0883-2691} \qquad Giovanna Castellano\orcidlink{0000-0002-6489-8628}\\
Department of Computer Science, University of Bari Aldo Moro, Italy\\
{\tt\small \{nicola.fanelli, gennaro.vessio, giovanna.castellano\}@uniba.it}
}

\twocolumn[{%
\renewcommand\twocolumn[1][t]{#1}%
\maketitle
\begin{center}
    \centering
    \captionsetup{type=figure}
\includegraphics[width=1.0\textwidth]{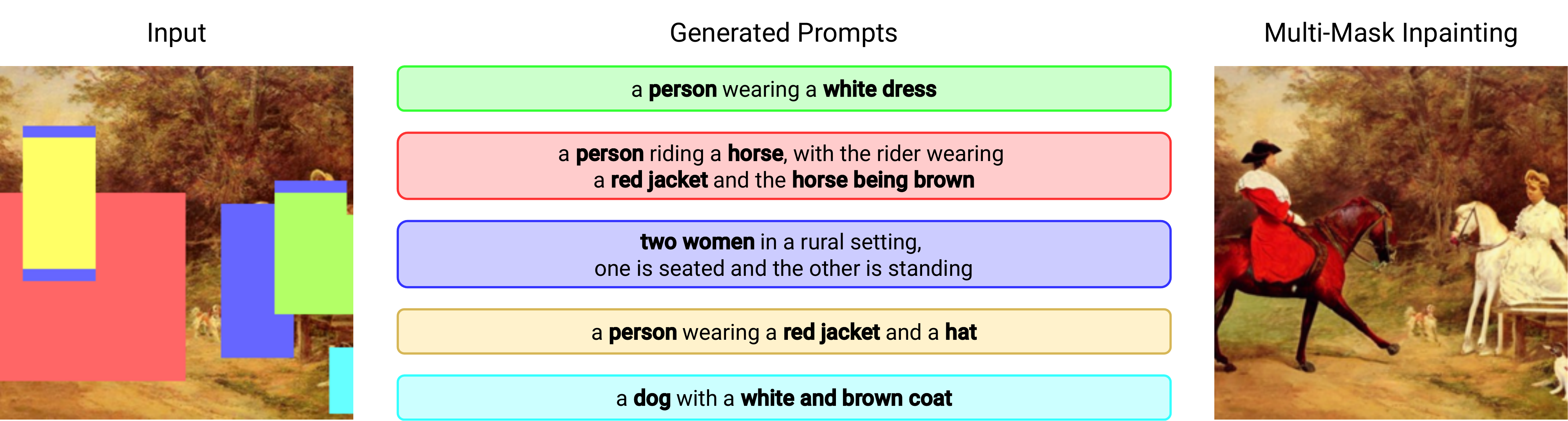}
    \captionof{figure}{We present a pipeline to address a novel task we refer to as \textit{text-guided multi-mask inpainting}. In this task, given an image with multiple masked regions, we aim to inpaint all regions simultaneously, with each region guided by its text prompt. Moreover, we demonstrate that it is possible to generate creative and plausible multi-mask text prompts automatically, starting solely from the masked image, thereby fully automating the inpainting process.}
\end{center}%
}]

\begin{abstract}
Inpainting focuses on filling missing or corrupted regions of an image to blend seamlessly with its surrounding content and style. While conditional diffusion models have proven effective for text-guided inpainting, we introduce the novel task of multi-mask inpainting, where multiple regions are simultaneously inpainted using distinct prompts. Furthermore, we design a fine-tuning procedure for multimodal LLMs, such as LLaVA, to generate multi-mask prompts automatically using corrupted images as inputs. These models can generate helpful and detailed prompt suggestions for filling the masked regions. The generated prompts are then fed to Stable Diffusion, which is fine-tuned for the multi-mask inpainting problem using rectified cross-attention, enforcing prompts onto their designated regions for filling. Experiments on digitized paintings from WikiArt and the Densely Captioned Images dataset demonstrate that our pipeline delivers creative and accurate inpainting results. Our code, data, and trained models are available at \href{https://cilabuniba.github.io/i-dream-my-painting}{https://cilabuniba.github.io/i-dream-my-painting}.
\end{abstract}

\section{Introduction}
\label{sec:intro}

Image inpainting involves restoring missing image regions by utilizing the surrounding context. Early approaches mainly focused on filling gaps with natural background patterns \cite{bertalmio2000image}. With the advent of deep learning, more advanced methods began incorporating semantic understanding along with global and local context \cite{pathak2016context,yeh2017semantic}. These techniques not only generate content for arbitrary regions but also make plausible predictions, such as inferring nostrils when the missing area is located below the nose.


Subsequent methods have incorporated additional semantic information, such as class categories, feature activations, shapes, or text \cite{miyato2018cgans,cai2022progressivegans,zeng2022shape,xie2023smartbrush,zhang2020text}, extending inpainting to full object generation. Text-to-image models like Stable Diffusion \cite{rombach2022high} have enabled text-guided inpainting, where users provide prompts to direct the generation process, allowing for creative outcomes (e.g., not just generating a nose, but a clown's nose). However, these models often default to probable objects or background textures with incomplete prompts and struggle with generating complex content without detailed guidance \cite{chiu2024brush2prompt}. Additionally, they are typically trained for inpainting using general image descriptions as text guidance for randomly selected image regions, which can cause text misalignment \cite{xie2023smartbrush}.

This paper proposes a new pipeline to address the mentioned challenges. First, we introduce a new task called \textit{text-guided multi-mask inpainting}, which involves inpainting multiple image regions simultaneously, each guided by a specific text prompt. We go further by exploring the automatic generation of textual prompts directly from a corrupted image, as done in Brush2Prompt \cite{chiu2024brush2prompt}, but with the added complexity of predicting multiple prompts for different regions in a single pass. We frame this as a visual instruction-following task for a multimodal large language model (MLLM) like LLaVA \cite{liu2024visual}. By fine-tuning this model using QLoRA \cite{dettmers2024qlora}, we automatically generate multiple prompts for use as user suggestions or direct text prompts for Stable Diffusion. Finally, we adapt the rectified cross-attention technique from FreestyleNet \cite{xue2023freestyle} and concatenate the multi-mask prompts into a single prompt, enforcing the correct parts of the prompt on their corresponding regions in the diffusion model, achieving similar quantitative results and more aesthetically pleasing qualitative outcomes compared to inpainting masks individually.


We focus our analysis on a dataset of digitized artworks from WikiArt, as these present a challenging application. Artworks often feature complex imagery and abstract concepts, demanding creativity for interpretation and identification, complicating prompt generation and text-guided inpainting. Since these images lack object-level annotations, we demonstrate that such annotations can be obtained using MLLMs like Kosmos-2 \cite{peng2024grounding} and LLaVA, showcasing our pipeline's applicability to datasets without annotations. To illustrate the method's adaptability to other domains, such as photographic scenes, we also test it on the Densely Captioned Images dataset \cite{urbanek2024picture}, which includes region-level annotations.


Our pipeline offers several benefits: it assists users of all skill levels with tools like Stable Diffusion, enables efficient multi-area inpainting with distinct prompts in a single pass, and serves as a data augmentation tool for computer vision tasks. In art, it can aid the analysis of damaged artworks by identifying potential missing objects in paintings.


\section{Related Work}
\label{sec:rel_work}

\paragraph{Diffusion Models} Diffusion models \cite{ho2020denoising} are state-of-the-art for image synthesis tasks \cite{dhariwal2021diffusion}. They use a forward process, where noise is added to an example over time via a Markovian or non-Markovian procedure \cite{song2021denoising}, and a backward process, where a neural network predicts and removes the noise to generate new samples. While initially designed to operate on raw image dimensions, recent approaches like Stable Diffusion \cite{rombach2022high} improve efficiency by working on encoded image representations from a variational autoencoder (VAE). Models conditioned on text, often through cross-attention, produce high-quality images but still struggle to fully adhere to prompts, prompting solutions like ControlNet \cite{zhang2023adding}, which adds extra conditioning inputs like images. Diffusion models have also been extended to inpainting in different ways \cite{pmlr-v162-nichol22a,rombach2022high,corneanu2024latentpaint}. Specifically, Stable Diffusion \cite{rombach2022high} achieves text-guided inpainting via further fine-tuning of the image synthesis model by adding the mask and the latent representation of the masked image as additional inputs, using zero-initialized weights and global image descriptions with masks randomly selected as in \cite{suvorov2022resolution}.

\paragraph{Multimodal Large Language Models}
Multimodal large language models extend large language models (LLMs) to handle multiple modalities, particularly vision, as both input and output, enabling a broad range of multimodal tasks \cite{dai2023instructblip,alayrac2022flamingo,liu2024visual}. In vision-language contexts, MLLMs can address tasks like image captioning and visual question answering and, more recently, solve various tasks through instruction-following prompts. Most MLLMs leverage CLIP \cite{radford2021learning} as the image encoder, benefiting from its strong image-text alignment and attempting to connect image features to an LLM for text generation. For example, LLaVA \cite{liu2024visual} employs a multimodal projector to incorporate image features alongside text tokens. Given their large size, these models, like LLMs, can be fine-tuned using parameter-efficient methods such as LoRA \cite{hu2022lora} or QLoRA \cite{dettmers2024qlora} for quantized models. In this paper, we fine-tune LLaVA using QLoRA for our specific task, as the base model did not yield satisfactory results in its original form.

\paragraph{Diverse Image Inpainting} Recent image inpainting approaches have aimed at generating more diverse outputs. For example, RePaint \cite{lugmayr2022repaint} utilizes an unconditional diffusion model to produce varied inpainting results from the same input. Cohen \etal \cite{cohen2024from} enhance diversity by sampling rare yet plausible concepts from the posterior distribution of generated objects. However, these methods do not leverage the additional guidance provided by text in conditional diffusion models. To address this, Brush2Prompt \cite{chiu2024brush2prompt} introduces a prompt generator that creates plausible and diverse prompts from corrupted images. This text-based approach empowers users to either use the inputs directly or as prompt suggestions, fully harnessing the capabilities of state-of-the-art conditional diffusion models. Similarly, we aim to generate prompts for a novel multi-mask setting, which adds complexity to the generation task.

\section{Our Approach}
\label{sec:ours}

\subsection{Data Annotation}
\label{sec:data_ann}

\begin{figure}[t]
  \centering
   \includegraphics[width=1.0\linewidth]{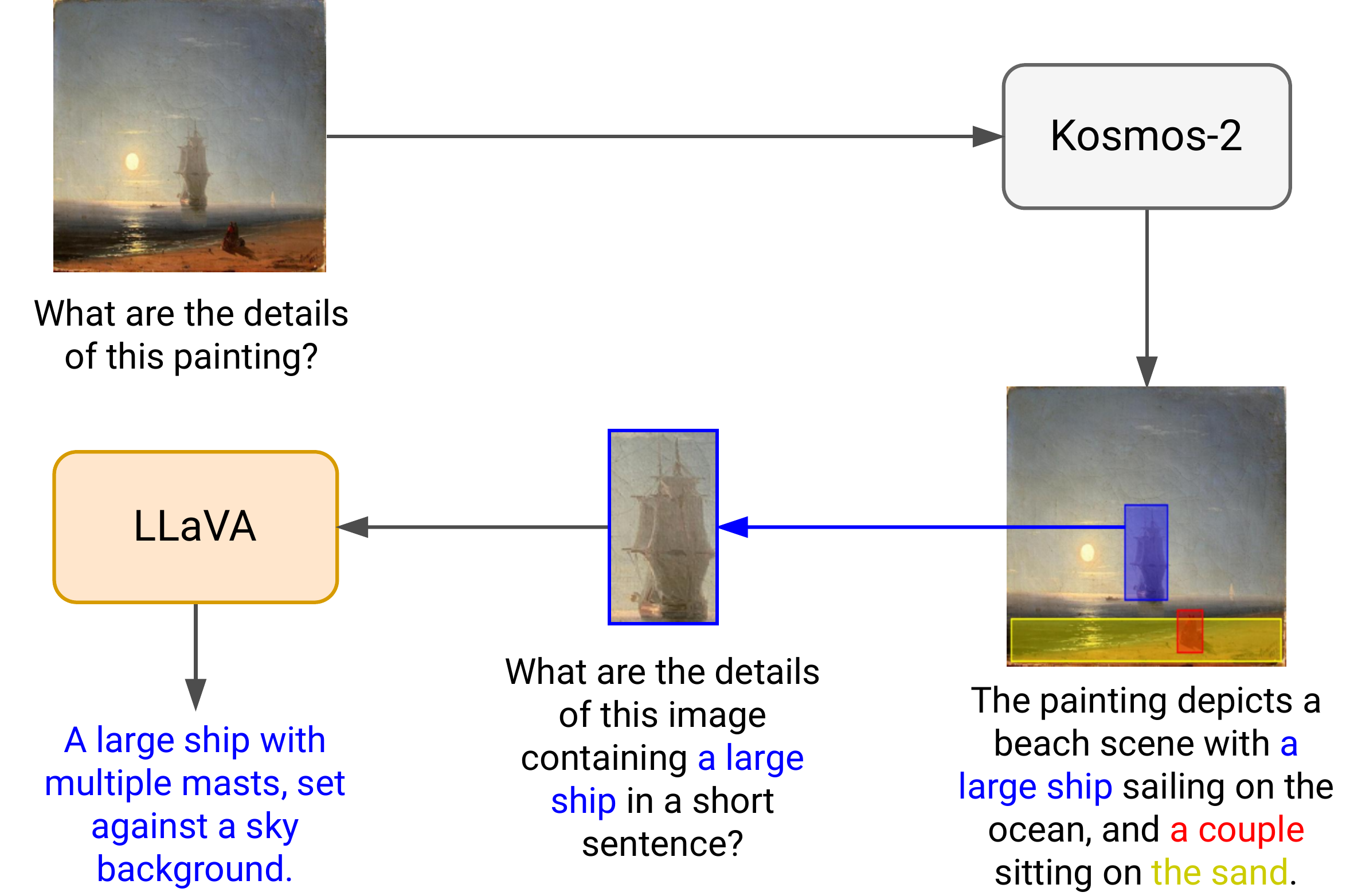}

   \caption{Overview of our automatic annotation process. We input artwork images into Kosmos-2 to obtain bounding box annotations for the main objects in each image. Afterward, we provide cropped images of these objects to LLaVA to generate more detailed object-level captions for our dataset.}
   \label{fig:data_annotation}
\end{figure}

Object-level annotations of images are essential to train both the prompt generator and the diffusion model for inpainting. To meet this requirement, we have developed a data annotation pipeline that leverages MLLMs to produce these annotations. The pipeline is organized into two primary steps, as illustrated in Fig.~\ref{fig:data_annotation}, described below.

\paragraph{Step 1: Identifying Objects and their Locations} In this initial step, our primary focus is identifying the objects of interest within the images. To accomplish this, we employ Kosmos-2, a model designed to provide grounded responses to user prompts. Specifically, we prompt Kosmos-2 to generate a grounded description of the image. The model responds with a concise caption, linking text tokens (usually noun chunks) to automatically generated bounding boxes that highlight the entities it identifies as most significant for inclusion in the image description.

\paragraph{Step 2: Obtaining Object-level Descriptions} In the second step, we require detailed descriptions at the object level, which are not provided directly by Kosmos-2. We first identify the bounding box associated with each grounded sequence of text tokens representing an entity to achieve this. We then crop the corresponding region of the image. Next, we prompt LLaVA to generate a detailed description of the object in the cropped image, enhancing the prompt by including the grounded text tokens (highlighted in blue in Fig.~\ref{fig:data_annotation}, right) to give LLaVA a contextual hint about the object. When an entity is associated with multiple bounding boxes, we crop the corresponding regions and create a collage by arranging these areas in a square grid. The grid cells are sized according to the largest width and height among the bounding boxes, and the cropped regions are placed in each cell. The entity is then treated as a single unit.

\subsection{Prompt Generation}

\begin{figure*}[t]
  \centering
   \includegraphics[width=1.0\textwidth]{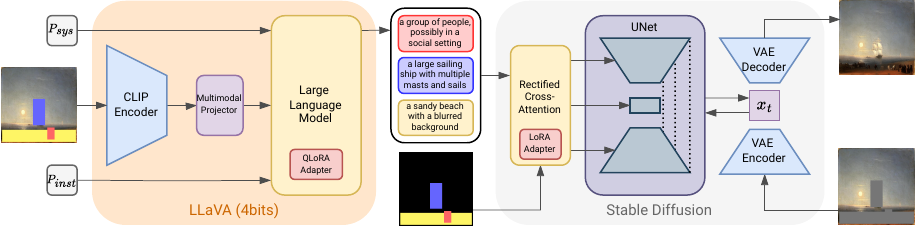}

   \caption{Overview of our pipeline. We utilize LLaVA as a prompt generator for multi-mask inpainting, integrating the generated multi-mask prompts and inpainting layout with Stable Diffusion through rectified cross-attention. The only trainable components of the pipeline are a QLoRA adapter for LLaVA, with the model quantized to 4 bits, along with a LoRA adapter that modifies the cross-attention layers of Stable Diffusion to accommodate the new task.}
   \label{fig:pipe}
\end{figure*}

We define the task of prompt generation for multi-mask inpainting as follows: given an image \( I \) with \( n \) corresponding binary masks \( \{M_i\}_{i=1}^{n} \), where \( 1 \leq n \leq N \), with $N$ being the maximum number of masks to be inpainted in a single example from the dataset, and \( n \) object-level textual descriptions \( \{Y_i\}_{i=1}^{n} \) that describe the entities corresponding to the masks, our goal is to generate texts \( \{\hat{Y}_i\}_{i=1}^{n} \), predicting what is hidden in the image behind each mask. This prediction must be made without access to the information in the image obscured by any of the masks. Mathematically, this is expressed as having access not to the entire $I$, but only to \( I \odot (1 - \bigcup_{i=1}^{n} M_i) \).

To address this problem, we fine-tune LLaVA, an MLLM designed for multimodal instruction-following tasks. Given an image $I$, a system prompt $P_{sys}$, an instruction $P_{inst}$, and a ground truth answer $A= a_1, \dots, a_T$, where $T$ is the number of tokens forming the answer, a single training example is defined as $S=(P_{sys}, I, P_{inst}, A)$. LLaVA is then trained to respond to instructions by minimizing the following multimodal causal language modeling loss with respect to the model parameters $\theta$:
\begin{equation}
    \mathcal{L}_{\theta}(S) = - \sum_{t=1}^{T} \log p_{\theta}(a_t | a_1, \dots, a_{t-1}, P_{sys}, I, P_{inst})
\label{eq:llava}
\end{equation}

Thus, for fine-tuning on our task, we prepare training examples from the dataset annotated as explained in Section~\ref{sec:data_ann} by adapting them to the previous formulation in the following manner:

\begin{itemize}
    \item First, since LLaVA accepts only a single image as input, we construct the input image $I_{inp}$ by assigning a random color to each mask $M_i$ associated with $I$ and overlaying them on $I$ with maximum opacity. Because the masks may overlap, we draw them from the largest to the smallest to minimize the possibility of obscuring any mask, and assign colors in random order.
    \item Second, we define the ground truth \( A \) by concatenating the ground truth object-level descriptions \( Y_i \) and enclosing each one within special tags that specify the corresponding color. For example, the ground truth for a two-mask example where the sampled colors are blue and red would be represented as $A = \texttt{\textcolor{blue}{<blue>}}, Y_1, \texttt{\textcolor{blue}{</blue>}}, \texttt{\textcolor{red}{<red>}}, Y_2, \texttt{\textcolor{red}{</red>}}$.
    \item Finally, we modify the corresponding instruction prompt $P_{inst}$ by including the names of the colors in the order they appear in $A$, along with a brief instruction describing the inpainting multi-prompt generation task.
\end{itemize}

\subsection{Multi-Mask Inpainting}

The ultimate goal of our pipeline is to inpaint the missing regions specified by the masks $M_i$ for $i=1, \dots, n$ using Stable Diffusion. This is done by inpainting each region according to the corresponding prompt $Y_i$ (or $\hat{Y}_i$, if the prompt has been generated). A key requirement for this process is that all regions are inpainted simultaneously, avoiding the need to repeat the process for each region, which would otherwise require $n$ separate passes through Stable Diffusion. However, Stable Diffusion for inpainting takes as inputs only the total mask $M = \bigcup_{i=1}^{n}M_i$ and the masked image $I \odot (1 - M)$, along with a single text prompt $Y$. By concatenating the object-level prompts $Y_i$ into a single prompt $Y_{inp}$, we lose the spatial correspondence between the prompts $Y_i$ and the masks $M_i$. This results in the model not adhering to the region assignments and, as demonstrated by our experiments, not fully understanding the prompt due to blending the subparts.

Therefore, we draw inspiration from the related task of freestyle layout-to-image synthesis (FLIS), which involves generating images based on an input layout where specific regions correspond to particular classes or text prompts, closely aligning with our problem. We adapt a technique from FLIS introduced by Xue \etal~in FreestyleNet \cite{xue2023freestyle}, known as rectified cross-attention (RCA). In diffusion models, prompt semantics are integrated into image synthesis via cross-attention layers, where image patches serve as queries $Q$ and text embeddings act as keys $K$ and values $V$, following the standard attention mechanism. This results in an attention map $\mathcal{A}$ calculated as:
\begin{equation}
    \mathcal{A} = \frac{QK^{\top}}{\sqrt{d}} \in \mathbb{R}^{T\times H \times W}
\end{equation}
where $T$ is the number of text embeddings obtained from the text encoding of the concatenated prompts $c_\phi({Y_1, \dots, Y_n})$, $H$ and $W$ are the height and width
of the latent image, and $\sqrt{d}$ is a normalization factor.

Recognizing that we can modify the logits of this attention map at the token level along the first dimension while maintaining spatial correspondences with the original image, we downsample the spatial layout to match the size of the latents. This layout is obtained by repeating the binary masks $M_i$ for each token in the corresponding prompt $Y_i$ and then stacking these repeated masks along the first dimension to create the layout $L \in \{0,1\}^{T \times H \times W}$. We then proceed to rectify the attention map by:
\begin{equation}
    \hat{\mathcal{A}}_{k,i,j}=\begin{cases}
    \mathcal{A}_{k,i,j} & \text{if } L_{k,i,j} = 1 \\
    -\inf & \text{otherwise}
    \end{cases}
\end{equation}

Thus, we adjust the tokens' weights for each prompt, making them more influential for the corresponding regions by utilizing the cross-attention mechanism.

Unlike FLIS, we adapt this technique by incorporating the following differences:
\begin{itemize}
    \item In our task, we have unmasked regions that do not correspond to any mask $M_i$. To address this, we modify each $M_i$ before incorporating it into the layout so that the unmasked regions attend to all tokens. Therefore, the mask for a single prompt $Y_i$ included in the layout is effectively $M_i=1 - \bigcup_{j=1,j\neq i}^{n}M_j$.
    \item In our task, intersections between masks to be inpainted may occur. We handle these naturally by representing the layout as a binary tensor, with separate channels for each mask, even though the original FLIS does not account for layouts with such intersections. This approach allows regions where multiple masks overlap to attend to the tokens of multiple prompts $Y_i$.
\end{itemize}


Figure~\ref{fig:pipe} provides an overview of our inference pipeline, where Stable Diffusion with RCA processes the generated prompts. Both tasks—prompt generation and multi-mask inpainting—only require training parameter-efficient adapters for the models.

\section{Experiments}
\label{sec:exp}

\subsection{Experimental Setup}

\subsubsection{Datasets}

We conducted experiments, using our entire pipeline, starting with data annotation on a dataset of digitized artistic paintings sourced from WikiArt. Specifically, we used images from WikiArt that correspond to artworks in $\mathcal{A}rt\mathcal{G}raph$ \cite{castellano2022leveraging}, a knowledge graph about art based on WikiArt and DBpedia. Notably, $\mathcal{A}rt\mathcal{G}raph$ contains 116,475 artworks by 2,501 artists, from which we collected images at the highest resolution available among three selected WikiArt thumbnail options.

We followed the data annotation procedure outlined in Sec.~\ref{sec:data_ann}, keeping in the dataset only bounding box annotations whose area was at least 1\% and at most 65\% of the total image area. We discarded images that did not have any bounding boxes meeting the size constraints, resulting in 102,276 images. We retained 5,000 validation images and 5,000 test images from the annotated set for our experiments.

In the subsequent steps of prompt generation and multi-mask inpainting, we retained up to five object-level annotations per image ($N=5$), selecting the largest ones. These annotations were then randomly shuffled and included in the training/testing examples, provided that the total area of the corresponding masks did not exceed 65\% of the image area. This approach did not significantly impact our dataset's overall distribution of mask quantities while ensuring that the examples used for inpainting contained a significant portion of unmasked content.

We performed this process for prompt generation on images at their original resolution, as LLaVA can handle different aspect ratios within its architecture. For multi-mask inpainting, we followed the same procedure, but first, we resized and center-cropped the images to a resolution of $512 \times 512$.

To assess our pipeline's effectiveness in use cases beyond artistic images, we also conducted experiments using the Densely Captioned Images (DCI) dataset \cite{urbanek2024picture}. This dataset consists of 7,805 photographic images sourced from SA-1B \cite{kirillov2023segment}, each accompanied by multiple region-caption pairs. We used the original annotations as ground truth, bypassing the automatic annotation process. Additional dataset details are available in the supplementary material.

\subsubsection{Implementation Details}

Our implementation leverages the Python libraries PyTorch and Hugging Face's Transformers for pre-trained vision-language models, Diffusers for diffusion models, and Accelerate for parallel training.

\paragraph{Data Annotation} We fed images into Kosmos-2 using the text prompt \textit{``What are the details of this painting?''} For LLaVA, we used the prompt \textit{``What are the details of this image containing \{noun chunk\} in a short sentence? Ignore the painting style,''} and let the model begin the generation with the prefix \textit{``The image shows.''} In the second step, we utilized LLaVA-1.6-Vicuna-13B \cite{liu2024llavanext}, with 4-bit quantization, which uses a 13B-parameter version of Vicuna \cite{vicuna2023} as the backbone LLM.

\paragraph{Prompt Generation} We used LLaVA-1.6-Vicuna-7B \cite{liu2024llavanext}, with 4bit quantization and a 7B-parameter backbone LLM, as the prompt generator. Specifically, we trained the model in mixed-precision using QLoRA, targeting all linear layers in the LLM with $r = 16$, $\alpha = 16$, and a dropout rate of $p = 0.05$, for a total of $\sim$40M trainable parameters. We employed AdamW with default settings as the optimizer and a constant learning rate schedule with a warmup phase for 1\% of the training steps, gradually increasing the learning rate to $\eta = 2 \cdot 10^{-4}$. Additionally, we clipped gradients to a maximum norm of 0.5. The model was trained for a single epoch (6 epochs on DCI) on four NVIDIA A100 GPUs over approximately 9.5 hours, with a total batch size of 32.

\paragraph{Multi-Mask Inpainting} We utilized Stable Diffusion-2.0-Inpainting \cite{rombach2022high} as our inpainting model. The model was trained in mixed precision using LoRA, specifically targeting all linear layers in the cross-attention blocks, for a total of $\sim$3M trainable parameters. We applied the same LoRA settings, optimizer, and learning rate schedule as those used for the prompt generation, except for the final learning rate, which was set to $\eta = 10^{-4}$. Gradients were clipped to a maximum norm of $1$. To maintain classifier free-guidance \cite{ho2021classifierfree}, we dropped text conditioning with a probability of $p = 0.1$, but only for the single-mask training examples. The model was trained to minimize the usual MSE objective of the diffusion model over uniformly sampled timesteps ($T=1000$) with a DDPM sampler \cite{ho2020denoising} for a single epoch (8 epochs on DCI) on one NVIDIA A100 GPU over approximately 1.5 hours, with a batch size of 32. At test time, we sampled images with 50 diffusion steps using the PNDM sampler \cite{liu2022pseudo} and a classifier free-guidance weight of 7.5.


\begin{table*}[t]
\centering{\small{
\begin{tabular}{@{}lccccc@{}}
\toprule
                         & Accuracy (\%) & BLEU@1 & BLEU@4 & ROUGE-L & CLIPSim \\ \midrule
LLaVA-Prompt                    & 7.74          & 20.81  & 1.30   & 19.99   & 22.46   \\
LLaVA-1Mask              & \textbf{36.52}         & 36.99  & 12.58  & {\ul 34.64}   & {\ul 24.65}   \\
\midrule
LLaVA-MultiMask-1Pred    & {\ul 35.48}             & \textbf{37.68}       & \textbf{13.15}       & \textbf{34.98}         & \textbf{24.79}       \\
LLaVA-MultiMask-LastPred & 33.08         & {\ul 37.40}  & {\ul 12.61}  & 34.45   & 24.46   \\
LLaVA-MultiMask-All      & 31.73         & 37.33  & 12.43  & 34.33   & 24.24 \\ \bottomrule
\end{tabular}
}}
\caption{Prompt generation quality results. We compared our LLaVA-MultiMask approach with the out-of-the-box LLaVA model and a fine-tuned single-mask version. 
The top 2 results are outlined in \textbf{bold} and with {\ul underline}.}
\label{tab:promptgen}
\end{table*}

\begin{table}[t]
\centering{\small{
\begin{tabular}{@{}lccc@{}}
\toprule
                         & A (\%) & B@1 & C \\ \midrule
LLaVA-Prompt                    & 9.52 & 1.79  & \textbf{24.15} \\
\midrule
LLaVA-MultiMask-LastPred & {\ul 31.43}         & 16.38  & 23.49 \\
LLaVA-MultiMask-All      & 29.91         & {\ul 18.22}  & 23.70 \\
LLaVA-MultiMask-All-Temp0.5 & \textbf{34.14} & \textbf{18.90} & \textbf{24.15} \\ 
\bottomrule
\end{tabular}
}}
\caption{Prompt generation quality results on the DCI dataset (A: Accuracy; B@1: BLEU@1; C: CLIPSim).}
\label{tab:promptgendci}
\end{table}

\begin{table*}[t]
\centering{\small{
\begin{tabular}{@{}lcccccc@{}}
\toprule
                       & FID \textdownarrow           & LPIPS \textdownarrow          & PSNR \textuparrow           & CLIP-IQA \textuparrow       & CLIPSim-I2I \textuparrow & CLIPSim-T2I \textuparrow \\ \midrule
SD-2-Inp-HQPrompt      & 19.18 (31.86)          & 22.82 (24.29)          & 14.55 ({\ul 14.36})          & 71.51 (73.89)          & 84.87 (85.63)           & 21.10 (20.92)            \\
SD-2-Inp               & {\ul 15.07} (\textbf{27.40}) & \textbf{21.90} (\textbf{23.63}) & {\ul 14.66} (14.35) & 73.10 (75.74)         & 88.87 (88.93)          & 25.70 (24.91)           \\
SD-2-Inp-RCA           & 15.39 (28.03)         & {\ul 21.98} (23.78)         & 14.59 (14.24)    & 73.24 (75.83)         & 88.83 (88.85)          & 25.81 (25.04)            \\
SD-2-Inp-FineTuned     & 15.49 (27.83)        & 22.06 (23.96)         & 14.44 (14.06)         & \textbf{74.64} (\textbf{77.68}) & 89.05 (89.04)     & 26.31 (25.40)      \\
\rowcolor{gray} SD-2-Inp-Repeated & \textbf{14.96} (27.56) & 22.47 (24.81) & \textbf{14.67} (\textbf{14.38}) & 70.58 (70.78) & {\ul 89.26} ({\ul 89.25}) & 25.81 (25.06)\\ 
\rowcolor{gray} SD-2-Inp-Repeated-FineTuned & 15.79 (29.20) & 22.73 (25.09) & 14.43 (14.11) & 72.29 (73.12) & 88.85 (88.91) & \textbf{26.95} (\textbf{26.23}) \\
\midrule
SD-2-Inp-RCA-FineTuned & 15.32 ({\ul 27.45})    & 22.00 ({\ul 23.74})    & 14.46 (14.13)          & {\ul 74.30} ({\ul 77.21})    & \textbf{89.28} (\textbf{89.35}) & {\ul 26.72} ({\ul 25.93})   \\ 
SD-2-Inp-RCA-FineTuned-Gen & 15.30 (27.94) & 22.69 (24.42) & 14.05 (13.64) & 72.80 (76.01) & 87.47 (87.68) & 23.25 (22.94) \\
\bottomrule
\end{tabular}
}}

\caption{Inpainting results. We compared our solution (SD-2-Inp-RCA-FineTuned), which leverages Stable Diffusion 2.0-Inpainting fine-tuned with RCA, against various non-fine-tuned and fine-tuned approaches. Results are reported for the full test set and for test examples with multiple masks (in parentheses). The gray rows show alternatives that repeat the inpainting process for each mask.}
\label{tab:inpainting}
\end{table*}

\subsubsection{Evaluation Metrics}

To assess the quality of the generated annotations, we used the CLIP cosine similarity (CLIPSim) between text embeddings and image embeddings. This reference-free metric has been empirically shown to correlate well with human judgments \cite{hessel2021clipscore} and is widely used for evaluating image-text alignment.

For prompt generation, we assessed how well the generated prompts match the real entities behind a mask to evaluate their plausibility. Accuracy was measured by checking if the root of the noun chunk from Kosmos-2 annotations appeared in the prompt. Additionally, we used text-matching metrics like BLEU \cite{papineni2002bleu} and ROUGE \cite{lin2004rouge}, and CLIPSim between the CLIP embeddings of the prompt and the corresponding mask crop for a reference-free measure. All metrics were averaged at the mask level. Intra-sentence diversity was measured with Distinct-N \cite{li-etal-2016-diversity}, and inter-sentence diversity with Self-BLEU \cite{zhu2018texygen}.

For multi-mask inpainting, we evaluated performance using standard metrics: Fréchet Inception Distance (FID) \cite{heusel2017gans}, Learned Perceptual Image Patch Similarity (LPIPS) \cite{zhang2018unreasonable}, and Peak Signal-to-Noise Ratio (PSNR). We also assessed image quality with the CLIP Image Quality Assessment score (CLIP-IQA) \cite{wang2023exploring} and measured prompt alignment using CLIP text-to-image (CLIPSim-T2I) and image-to-image similarity (CLIPSim-I2I). For the last two metrics, we focused on each specific masked region to be inpainted and its corresponding prompt. Following Lüddecke and Ecker's approach \cite{luddecke2022image}, we darkened and blurred the image background before inputting it into CLIP to improve local image-text alignment. A visualization of the computation of the metric is available in the supplementary material.

\begin{figure}[t]
  \centering
  \includegraphics[width=1.0\linewidth]{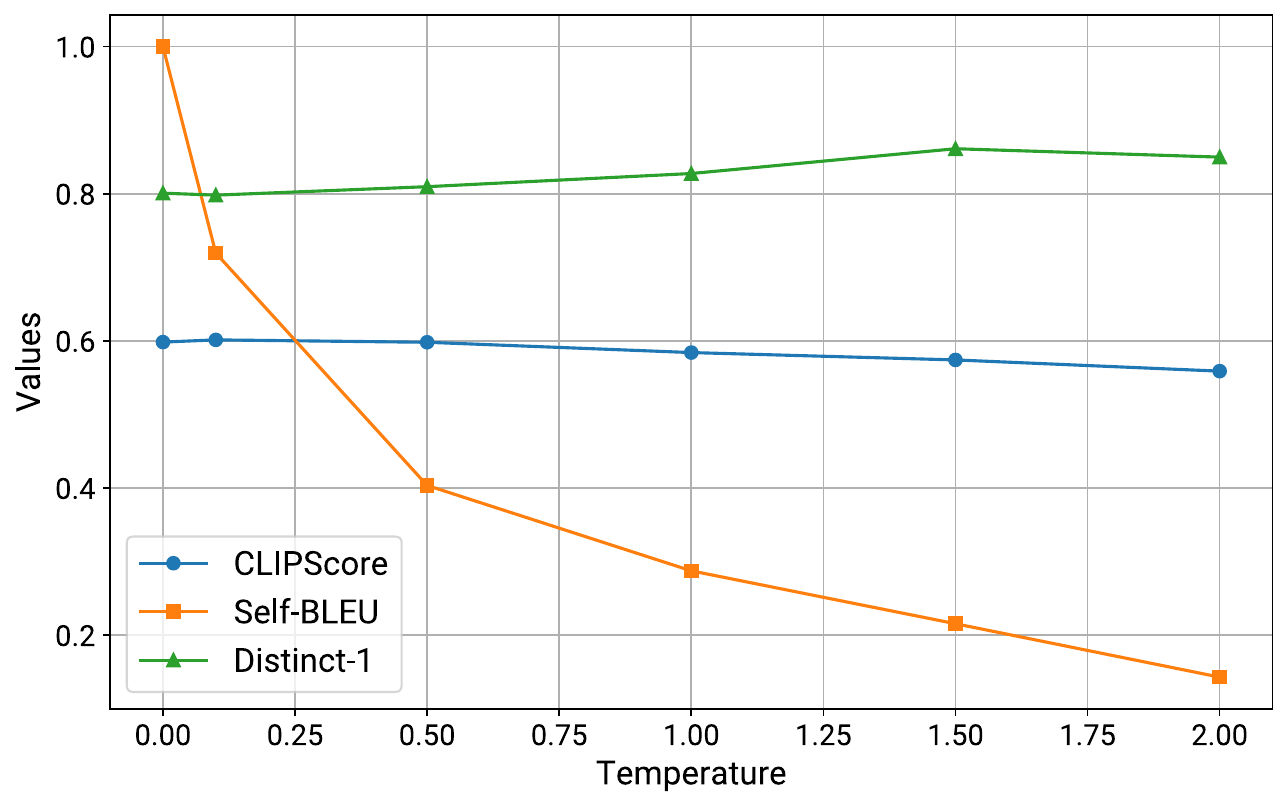}
  \caption{Effects of LLaVA sampling temperature on the quality and diversity of prompt generation. For CLIPSim, we report the similarity value scaled by 2.5 to map it into the $[0, 1]$ interval, allowing for easier comparison, as done in \cite{hessel2021clipscore}.}
  \label{fig:temperature}
\end{figure}

\subsection{Results and Discussion}

\paragraph{Data Annotation Quality} We computed the CLIPSim score between our art dataset images and the grounded descriptions generated by Kosmos-2 to assess the quality of object identification. We also calculated the similarity between bounding box crops and object-level descriptions generated by LLaVA. The average similarity scores are 0.32 and 0.28, respectively, indicating strong global and local alignment of the generated annotations. This is supported by empirical testing and human inspection while creating datasets like LAION-5B \cite{schuhmann2022laion}, which used a threshold of 0.28 to include image-text pairs.

\paragraph{Prompt Generation Quality and Plausibility} Table~\ref{tab:promptgen} compares our fine-tuning approach for LLaVA, which includes mask coloring and prompt-answer formatting, against the base LLaVA model's visual instruction-following abilities. The base model, limited to visual instruction-following, struggles with multi-mask prediction, often defaulting to general image descriptions. Even in a simplified single-mask scenario (using the Brush2Prompt paper's prompt), it inaccurately describes hidden objects, focusing on visible entities, leading to lower scores than our fine-tuned variants.

We fine-tuned two models: LLaVA-1Mask, trained on single-mask examples, and LLaVA-MultiMask, trained in a multi-mask setting. LLaVA-1Mask performs well in single-mask predictions, but LLaVA-MultiMask surprisingly outperforms it, even in single-mask scenarios (1Pred), suggesting that multi-mask training enhances prompt generation capabilities. Additionally, we evaluated LLaVA-MultiMask in a full multi-mask test, focusing on a single selected mask (LLaVA-MultiMask-LastPred). Despite having access to 18\% fewer unmasked regions on average, performance degradation was minimal, underscoring the robustness of the multi-mask model.

On DCI (Table~\ref{tab:promptgendci}), while the base model's prompts achieve CLIPSim scores comparable to LLaVA-MultiMask, they fall short in accuracy and BLEU. This likely stems from DCI's complex annotations, which are often truncated by the CLIP encoder, reducing CLIPSim for both models.


\paragraph{Temperature and Diversity} We investigated the impact of temperature adjustment on the quality and diversity of prompt generation. In this experiment, we applied sampling with a specified temperature to generate prompts four times for each of 128 randomly selected test examples (Fig.~\ref{fig:temperature}).

Our results indicate that increasing the temperature up to 0.5 results in only a minor reduction in CLIPSim while slightly boosting Distinct-1 and significantly lowering Self-BLEU, reflecting the generation of more varied prompts. This suggests that intermediate temperature values between 0.5 and 1 can offer users diverse prompt suggestions without compromising plausibility. 

Interestingly, increasing the temperature improves results on DCI (Table \ref{tab:promptgendci}), suggesting a positive correlation between diversity and plausibility in smaller datasets.

\begin{figure*}[t]
  \centering
   \includegraphics[width=1.0\textwidth]{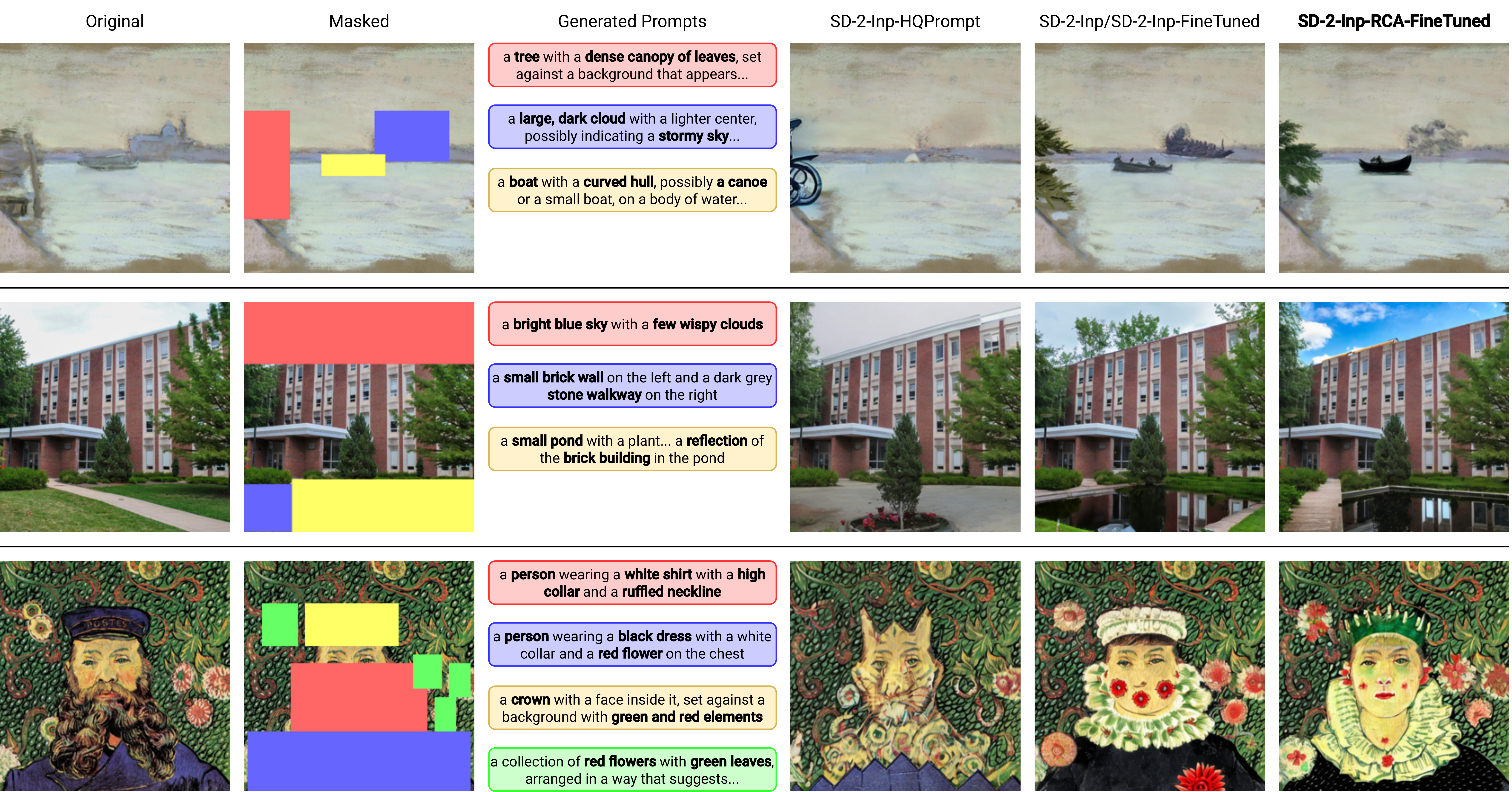}

   \caption{Qualitative results. We compare the different approaches tested in our evaluation. Fine-tuning with RCA enables the model to better follow prompts, capture details, and respect inpainting locations while reducing susceptibility to concept blending. 
   The second example is from DCI.}
   \label{fig:qual}
\end{figure*}

\begin{table}[t]

\centering{\small{
\begin{tabular}{@{}lccc@{}}
\toprule
                         & L \textdownarrow & P \textuparrow & C \textuparrow \\ \midrule
SD-2-Inp-HQPrompt                    & 28.81 & \textbf{12.95} & 22.81 \\
SD-2-Inp & \textbf{28.08}         & {\ul 12.79}  & {\ul 24.85} \\
\midrule
SD-2-Inp-RCA-FineTuned      & {\ul 28.64}         & 12.49 & \textbf{25.48} \\
SD-2-Inp-RCA-FineTuned-Gen & 30.26 & 12.16 & 23.66 \\ 
\bottomrule
\end{tabular}
}}
\caption{Inpainting results on the DCI dataset (L: LPIPS; P: PSNR; C: CLIPSim-T2I).}
\label{tab:inpaintingdci}
\end{table}

\paragraph{Inpainting Quality and Matching} We report multi-mask inpainting results in Table~\ref{tab:inpainting}. For this experiment we masked all the concepts to fill and provided the layout in input if the model supports RCA based on our method, using the ground truth prompts.

We provide two baselines where we tried SD-2-Inpainting as-is with the general prompt \textit{``High quality''} (SD-2-Inp-HQPrompt) or with the multi-mask prompt (SD-2-Inp). We notice that the second model achieves the best FID, LPIPS, and PSNR at the cost of not closely adhering to the prompts, as shown by the CLIPSim scores. After further inspection, we noticed that the SD-2-Inp 
model tends to blend all the concepts in a single prompt, sometimes filling specific masks with the background or unrecognizable objects. Thus, evaluating only the first three metrics is unsuitable for this task.

We applied RCA to the model without fine-tuning, observing a slight improvement in prompt-following, indicating its potential value for our task. We then fine-tuned two versions on the new dataset: one without RCA (SD-2-Inp-FineTuned) and one with RCA (SD-2-Inp-RCA-FineTuned). The RCA-enhanced model consistently improves prompt-following, with RCA specifying object regions more effectively. Our final model achieves the best prompt-following scores, especially in multi-mask scenarios, and ranks second in LPIPS, CLIP-IQA, and multi-mask FID among the tested options. 

We also include inpainting results obtained by repeating the object generation for each mask for completeness. These alternatives benefit from untruncated prompts and precise knowledge of the inpainting locations, giving them an advantage over other approaches. Despite this, our results show that the RCA approach achieves comparable outcomes while significantly improving efficiency by completing the inpainting process in a single step. Further analysis of generation across multiple mask settings is in the supplementary material.

Finally, we evaluated the entire pipeline using our fine-tuned Stable Diffusion with RCA and generated prompts. As expected, this model achieves the second-best FID score and outperforms SD-2-Inp-HQPrompt in inpainting ground truth entities, showing significant improvement even without access to ground truth prompts.

The improved region-prompt alignment from RCA is confirmed by results on DCI (Table \ref{tab:inpaintingdci}), where the RCA variant of the model outperforms the alternatives in CLIPSim.

\paragraph{Qualitative Evaluation} We present qualitative results of the entire pipeline, from prompt generation to multi-mask inpainting, in Fig.~\ref{fig:qual}. Despite the noise in the initial object-level annotations, the prompt generator produces plausible prompts but sometimes repeats concepts for overlapping masks or generates overly long prompts with unnecessary details. Improving this could involve NLP techniques for preprocessing annotations or postprocessing prompts. With RCA fine-tuning, the model follows prompts with more precision, as seen in the second example, where it captures the details of the sky. It is also less prone to concept blending: in the first example, the model without RCA confuses a cloud with a boat, and in the third, it blends prompts for flowers and ignores the crown. Multi-mask inpainting still struggles in crowded areas with overlapping or tiny masks, where prompts may be ignored. More qualitative results are provided in the supplementary material.

\section{Conclusion}
\label{sec:concl}

This paper introduced a new task: inpainting multiple image regions, each guided by its own text prompt. We proposed a pipeline to address this task, including automatic multi-mask prompt generation for full automation. For inpainting, we focused on enabling the model to interpret long, complex prompts, using RCA from FLIS to enforce prompt-region alignment. 

Our experiments evaluated prompt plausibility, diversity, and inpainting quality. We demonstrated that multi-mask inpainting with generated prompts is feasible, testing on both digitized paintings and photographic images. This pipeline has potential for various applications, but as discussed in the experimental section, some limitations warrant further research.

\paragraph{Acknowledgement} We acknowledge the CINECA award under the ISCRA initiative, 
which gave us access to computing resources and support. Nicola Fanelli's research is funded by a Ph.D.~fellowship under the Italian ``D.M.~n.~118/23'' (NRRP, Mission 4, Investment 4.1, CUP H91I23000690007).


%% file: Suppl.tex
\section{Additional Dataset Information}
\label{sec:dataset}

This section provides additional details on the automatically annotated dataset of digitized images of artworks from $\mathcal{A}rt\mathcal{G}raph$. As described in the main paper, the images were downloaded as thumbnails from the WikiArt API, using the formats \textit{HalfHD}, \textit{Large}, and \textit{PinterestLarge}. For each image, we selected the highest resolution available. The counts and resolutions of the images are summarized in Table \ref{tab:resolutions}.

\begin{table}[b]
\centering{\small{
\begin{tabular}{@{}lccc@{}}
\toprule
Thumbnail name & Width & Height (max val.) & No. of images \\
\midrule
HalfHD & 1366 & 800 & 58744 \\
Large & 750 & 600 & 31841 \\
PinterestLarge & 280 & 1120 & 25890 \\
\bottomrule
\end{tabular}
}}
\caption{Image thumbnail specifics for the 116,475 artwork in $\mathcal{A}rt\mathcal{G}raph$, downloaded from the WikiArt API.}
\label{tab:resolutions}
\end{table}

\begin{figure*}[t]
  \centering
   \includegraphics[width=1.0\textwidth]{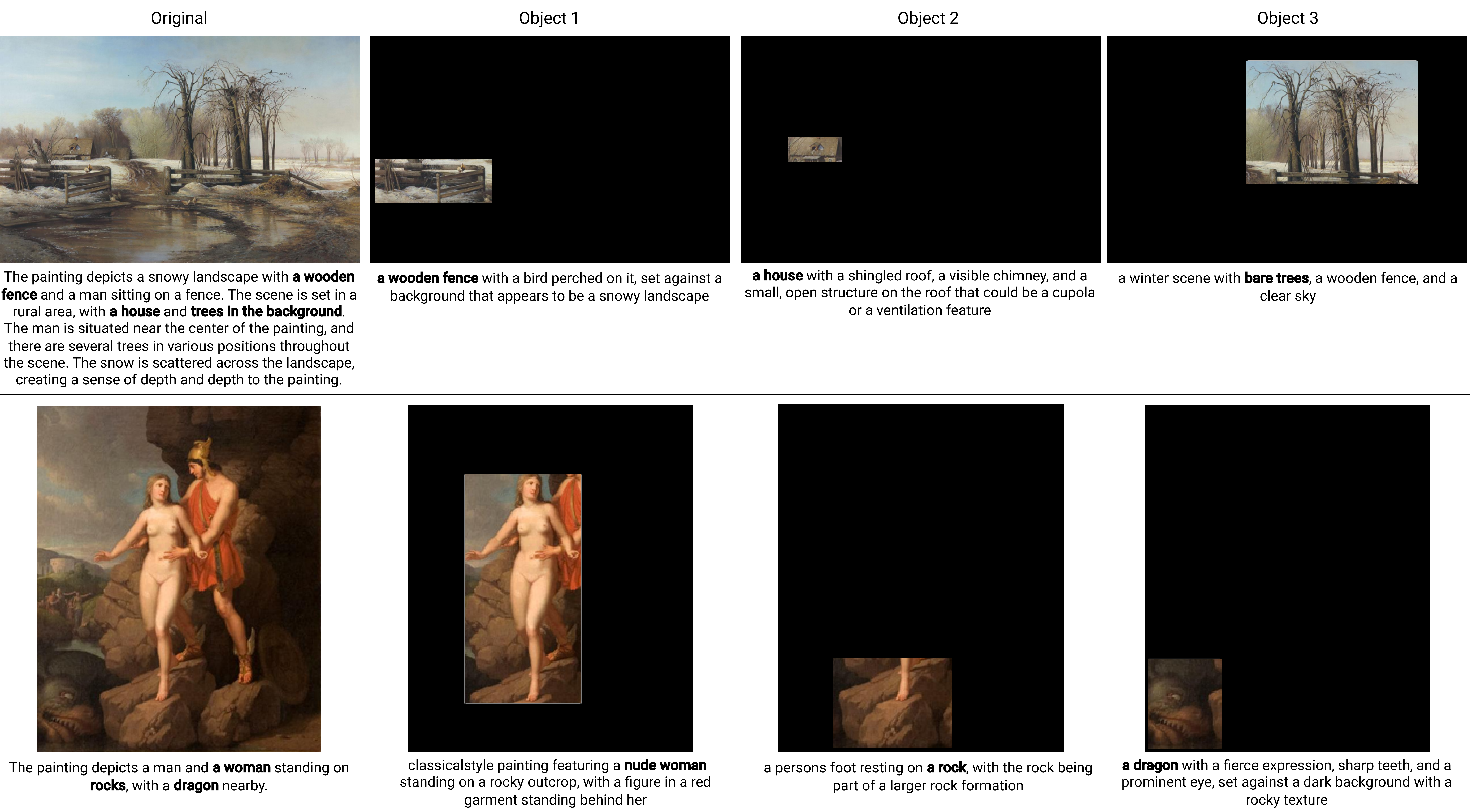}

   \caption{Automatically annotated samples from the art dataset. The first column displays the original images, along with the global grounded descriptions produced by Kosmos-2. Objects of valid sizes are shown in the other columns, with the rest of the image masked. These objects were cropped and provided to LLaVA to obtain the displayed object-level descriptions, which we used as training prompts. As demonstrated in the second example, certain masks can overlap. Our RCA implementation allows the intersection areas to attend to the prompts of both masks. Both samples present three valid masks.}
   \label{fig:data}
\end{figure*}

All 116,475 downloaded images were processed with Kosmos-2, generating grounded descriptions with bounding boxes around the primary objects depicted. After applying the size constraints detailed in the main paper, the dataset was refined to 102,276 images. We then used LLaVA to generate more detailed object-level descriptions for each object, limiting the descriptions to 40 tokens per object. Figure \ref{fig:data} illustrates two examples of images annotated using Kosmos-2 and LLaVA. As illustrated in the samples, automatically generated descriptions are inherently error-prone. However, in the main paper, we assessed their quality using CLIP, indicating that the alignment between the images and the generated texts remains acceptable. Caption errors may have had a greater impact on training the multi-mask inpainting diffusion model than on the MLLM used for prompt generation. This is because the original objects in the image are not directly involved in calculating the causal language modeling loss, which relies only on the text and the corrupted image input.

\begin{figure*}[t]
    \centering
    \begin{subfigure}[b]{0.46\textwidth}
        \centering
        \includegraphics[width=1.0\linewidth]{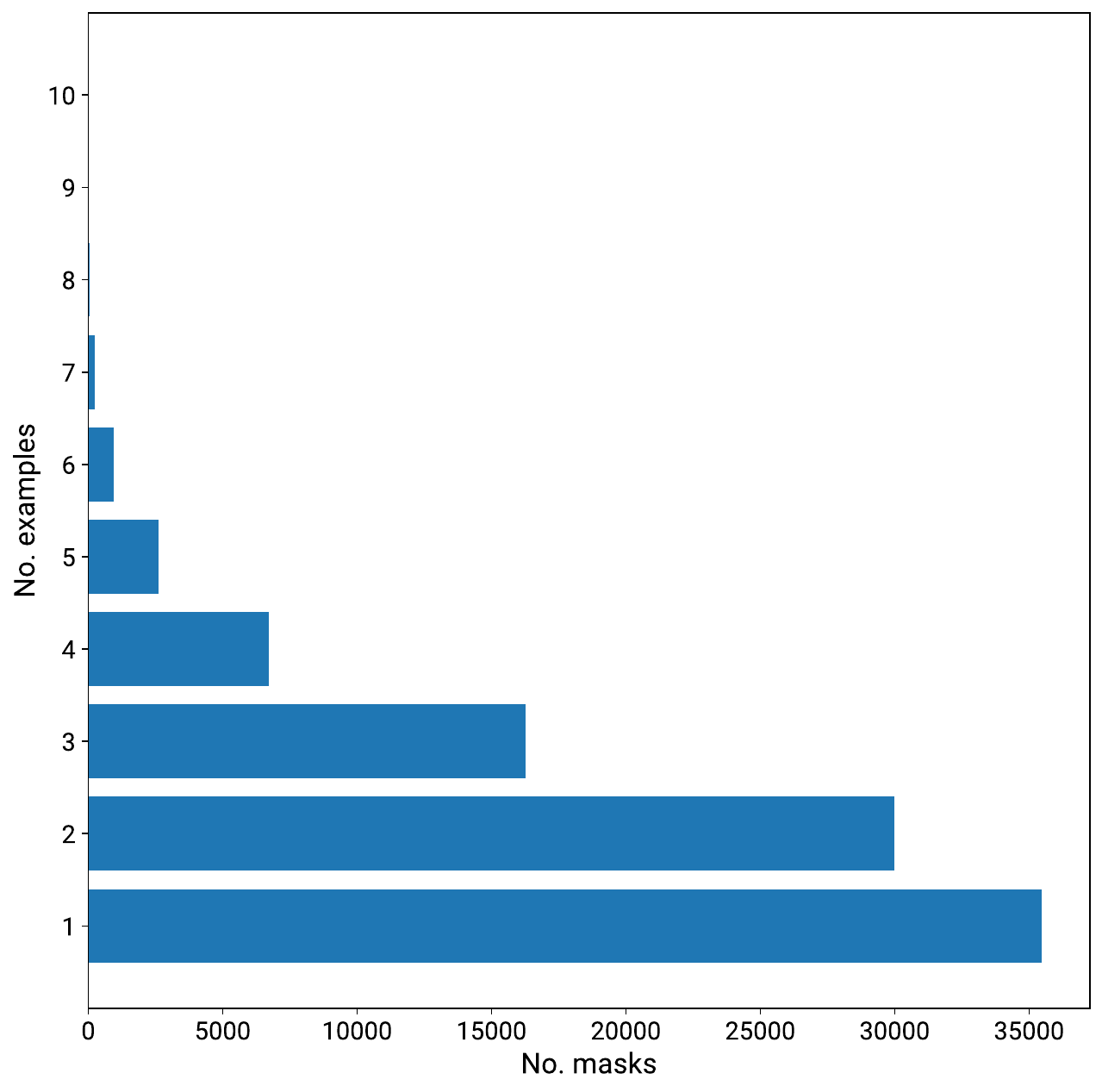}
        \caption{No.~of masks per example.}
        \label{fig:before}
    \end{subfigure}
    \hfill
    \begin{subfigure}[b]{0.46\textwidth}
        \centering
        \includegraphics[width=1.0\linewidth]{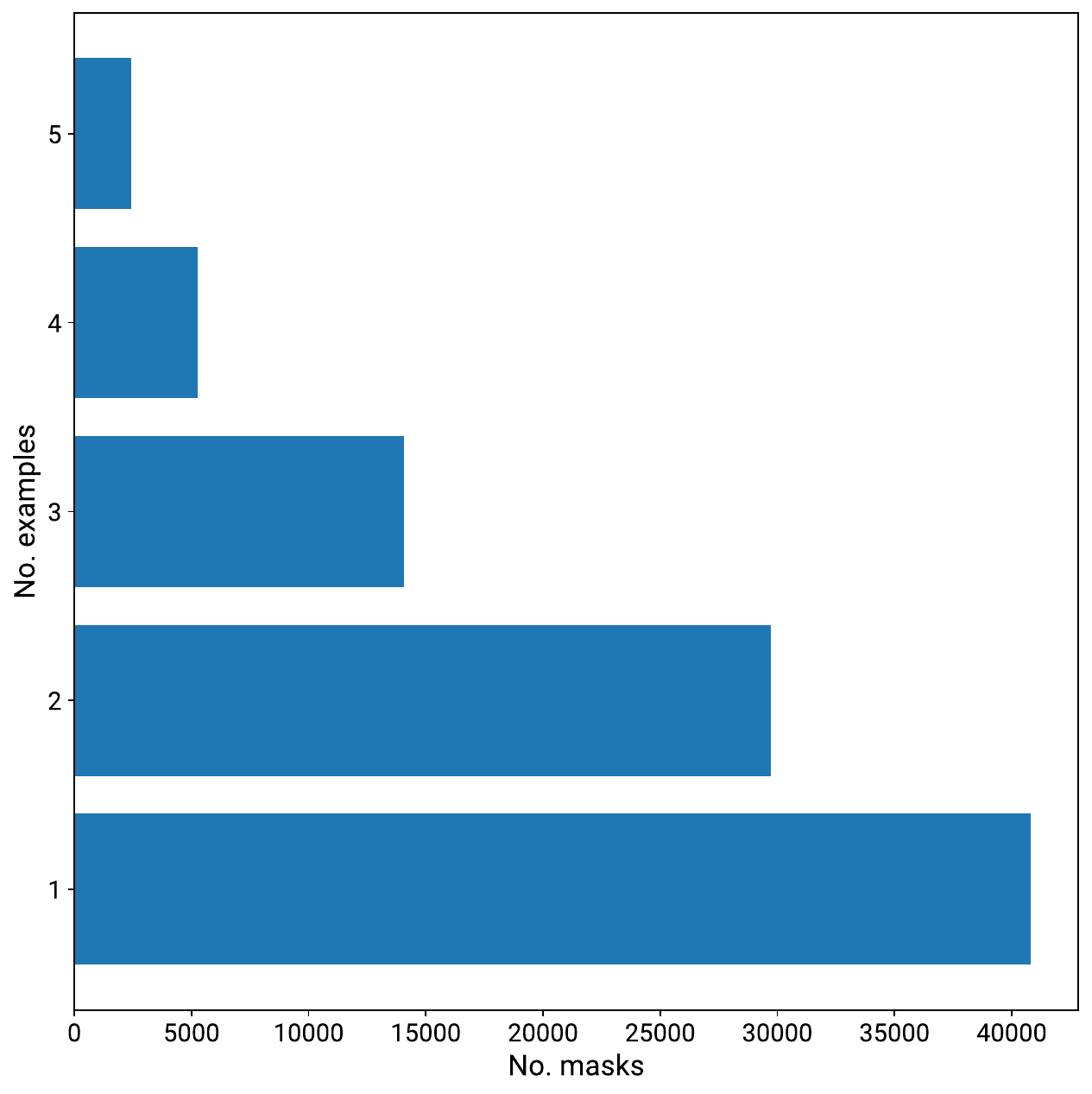}
        \caption{No.~of masks per example for training and testing.}
        \label{fig:after}
    \end{subfigure}
    
    \vspace{1em}  
    
    \begin{subfigure}[b]{0.46\textwidth}
        \centering
        \includegraphics[width=1.0\linewidth]{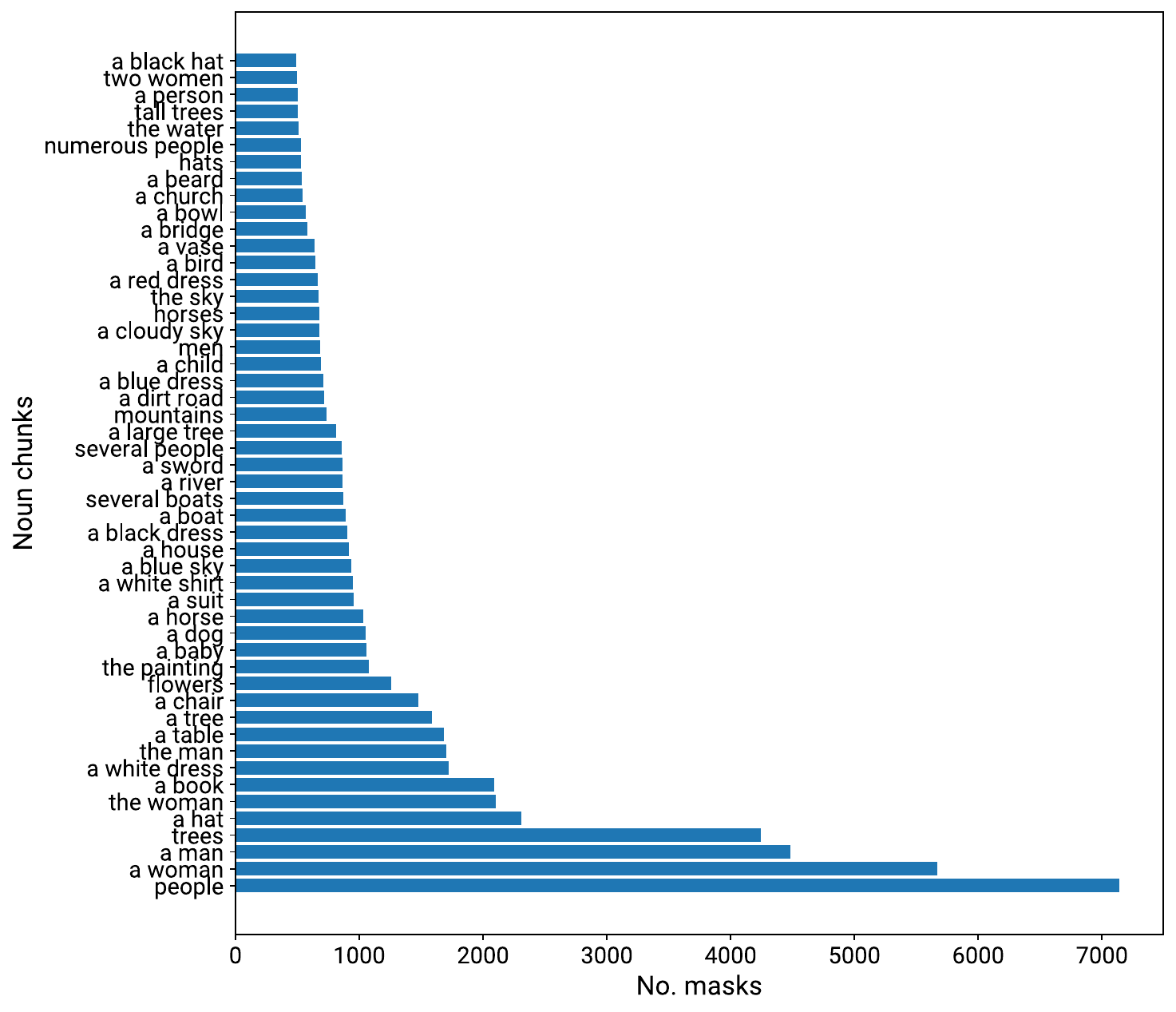}
        \caption{Most common noun chunks.}
        \label{fig:noun_chunks}
    \end{subfigure}
    \hfill
    \begin{subfigure}[b]{0.46\textwidth}
        \centering
        \includegraphics[width=1.0\linewidth]{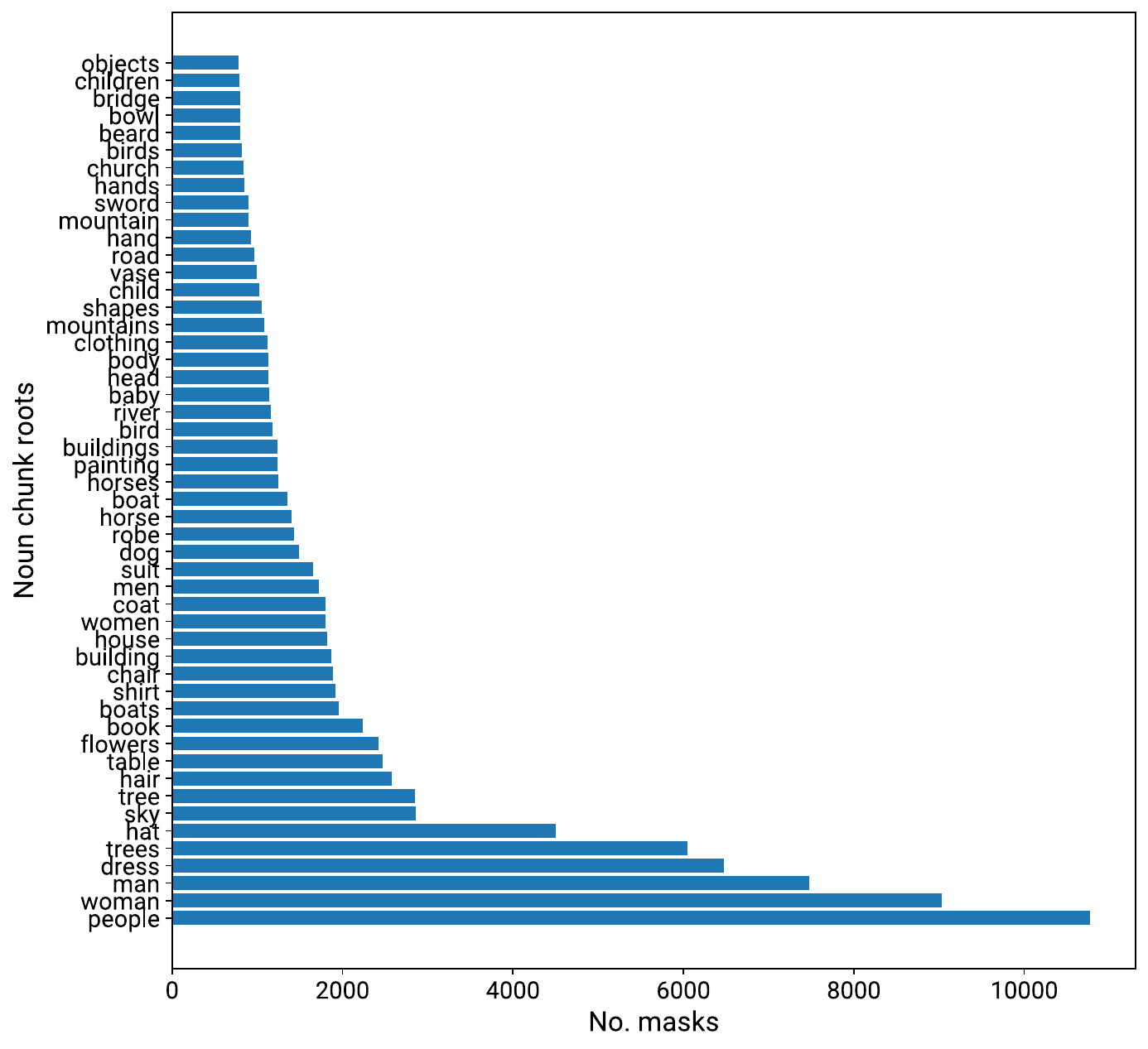}
        \caption{Most common noun chunk roots.}
        \label{fig:noun_chunk_roots}
    \end{subfigure}
    
    \caption{Dataset statistics on the number of masks per example and the objects depicted in the masks (noun chunks).}
    \label{fig:combined}
\end{figure*}

Additionally, we analyze the distribution of the number of masks per image both before (Fig.~\ref{fig:before}) and after (Fig.~\ref{fig:after}) applying a maximum limit of five objects for multi-mask inpainting, along with a global threshold on the covered area. We also present the distributions of the 50 most common objects, categorized by their respective noun chunks (Fig.~\ref{fig:noun_chunks}) and noun chunk roots (Fig.~\ref{fig:noun_chunk_roots}), which were used to compute the prompt generation accuracy. These distributions reveal two long-tailed patterns, with a strong skew towards common concepts.

\section{Additional Experiments}

In this section, we present additional experiments that were not discussed in the main paper.

\begin{figure}[t]
  \centering
   \includegraphics[width=1.0\linewidth]{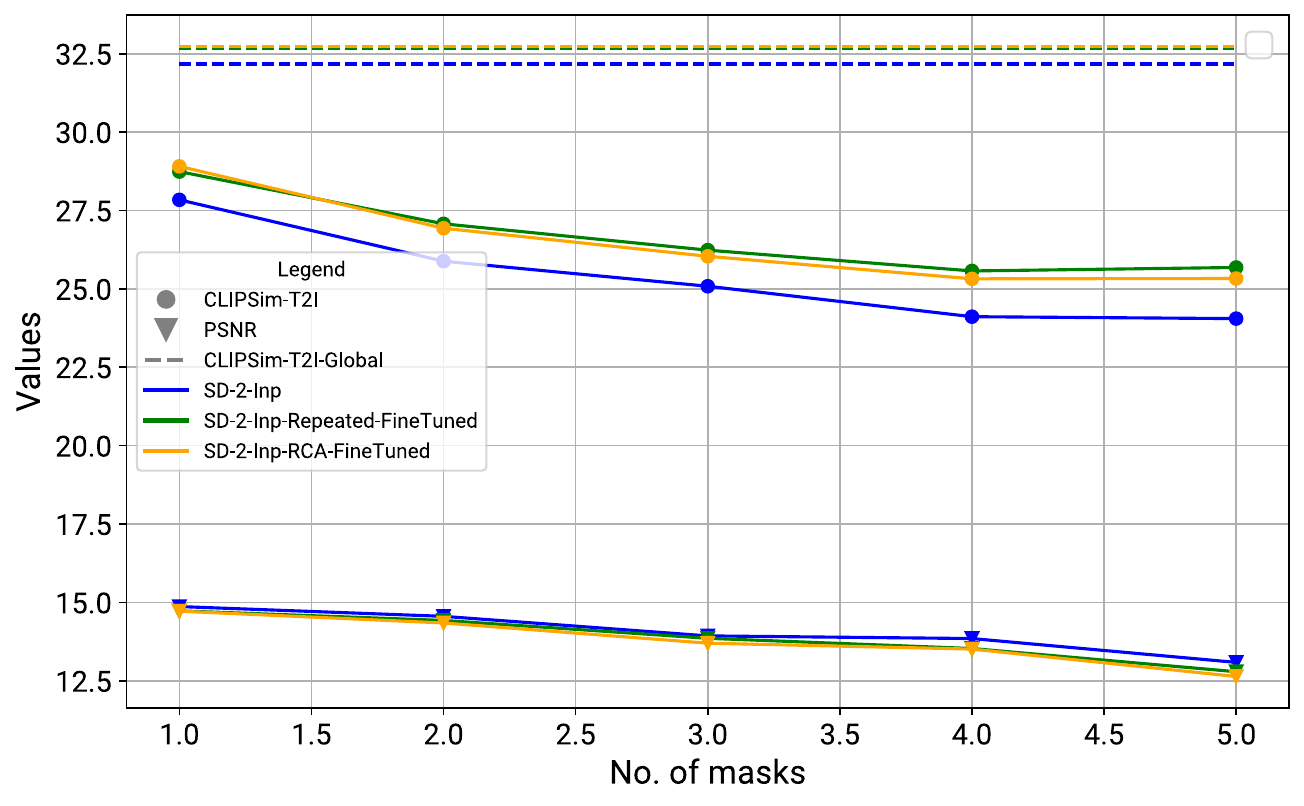}

   \caption{Comparative analysis on multi-mask inpainting.}
   \label{fig:exp1}
\end{figure}

\subsection{Analysis on the Number of Masks}


We compare the base Stable Diffusion-2.0-Inpainting model, a fine-tuned version that repeats inpainting for each mask, and our RCA fine-tuned version, evaluating performance across varying numbers of masks using ground-truth prompts truncated to a maximum of 10 words (Fig.~\ref{fig:exp1}).

The figure shows that our method consistently achieves better region-prompt alignment than the base Stable Diffusion model, regardless of the number of masks. Notably, as the number of masks increases, prompt-following scores decline, with the largest drop occurring when transitioning from single-mask to multi-mask inpainting, highlighting the increased difficulty of the task.

Interestingly, RCA slightly outperforms the repeated inpainting variant in the single-mask setting, while performing similarly in settings with more masks. This is significant because the RCA model requires a single backward pass and does not benefit from limiting the inpainting region to a single prompt. Both methods outperform the base Stable Diffusion-2.0-Inpainting model.

We also report the overall CLIPSim-T2I scores based on global image descriptions from Kosmos-2 during annotation. Both fine-tuned variants exceed the baseline, with RCA slightly outperforming the repeated inpainting approach.


Regarding PSNR values, the base Stable Diffusion model consistently scores slightly higher. However, qualitative assessments do not support the superiority of SD-2-Inp over RCA, indicating that PSNR can be influenced by various factors—such as the base model's tendency to fill masked regions with common or background elements—potentially skewing its evaluation.

\begin{figure}[t]
  \centering
   \includegraphics[width=1.0\linewidth]{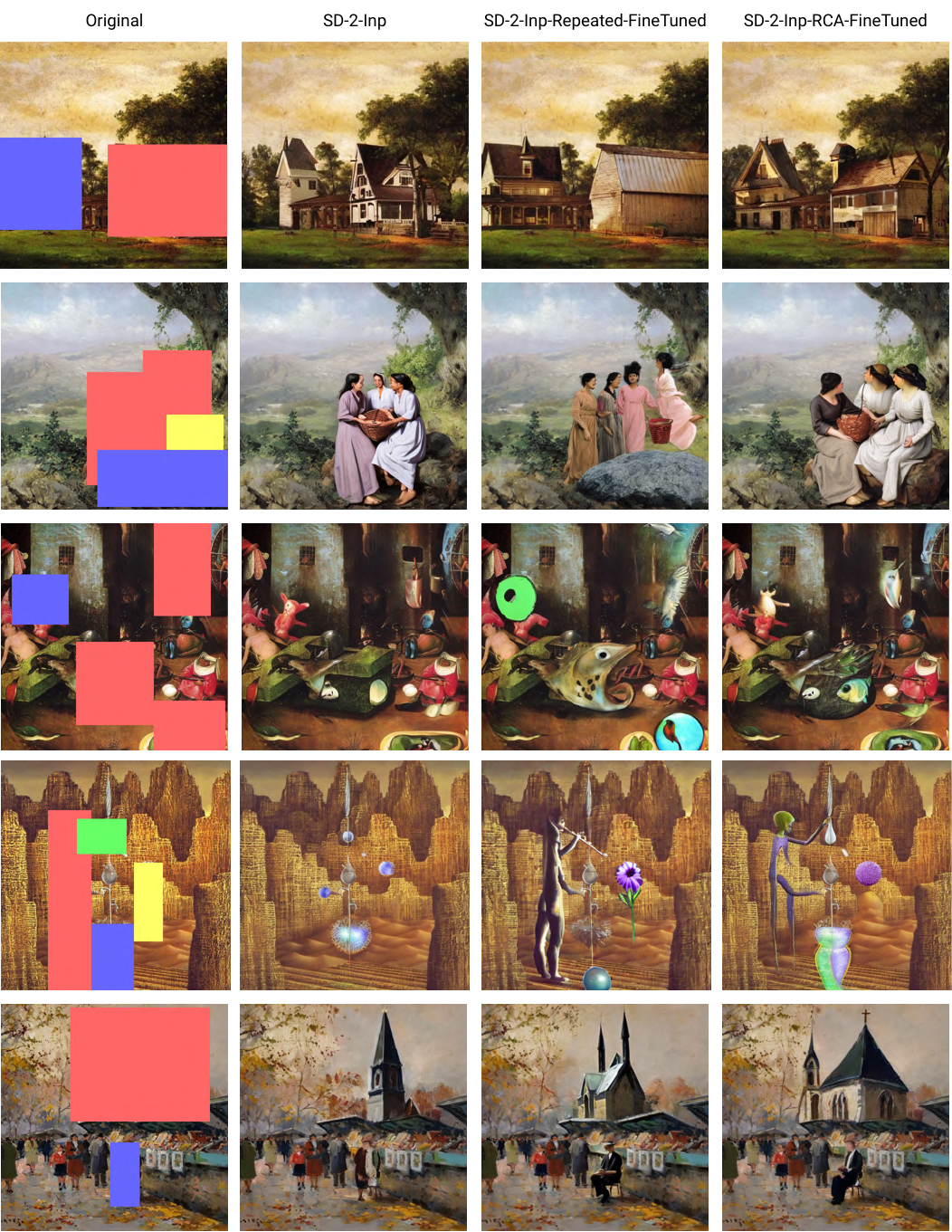}

   \caption{Qualitative comparison between repeated mask inpainting and multi-mask inpainting with RCA.}
   \label{fig:exp11}
\end{figure}

A qualitative comparison is shown in Fig.~\ref{fig:exp11}. Analyzing the generated images reveals that while repeating the inpainting process for each mask improves region-prompt alignment, it negatively impacts the overall aesthetics of the images. This approach sometimes introduces objects into the scene without meaningful integration with the surrounding elements. This issue is particularly evident in the second row of examples, where RCA prioritizes creating a cohesive scene by positioning the basket to interact meaningfully with its surroundings, even if this comes at the expense of precise positioning. In contrast, the repeated inpainting method appears to ``paste" the basket into the scene without establishing such connections.

Moreover, images produced with repeated inpainting exhibit reduced stylistic coherence and are more prone to unintended modifications in regions that should remain untouched. These observations highlight the limitations of relying solely on numerical metrics to evaluate inpainting models. While such metrics provide a baseline assessment of performance, they often fail to capture these nuanced aspects, which are especially significant in the context of artistic imagery.

\subsection{Domain Transfer}

\begin{table}[t]
\centering{\small{
\begin{tabular}{@{}lcccc@{}}
\toprule
& L \textdownarrow & P \textuparrow & Q \textuparrow & C \textuparrow \\ \midrule
DCI pipeline & \textbf{30.26} & 12.16 & 74.56 & \textbf{23.66} \\
Art pipeline & 30.69 & \textbf{12.26} & \textbf{80.25} & 23.00 \\
\bottomrule
\end{tabular}
}}
\caption{Comparative results on DCI between the dataset-specific pipeline and the pipeline transferred from the art domain (P: PSNR; L: LPIPS; Q: CLIP-IQA; C: CLIPSim-T2I).}
\label{tab:domain_transfer}
\end{table}

\begin{figure}[t]
  \centering
   \includegraphics[width=1.0\linewidth]{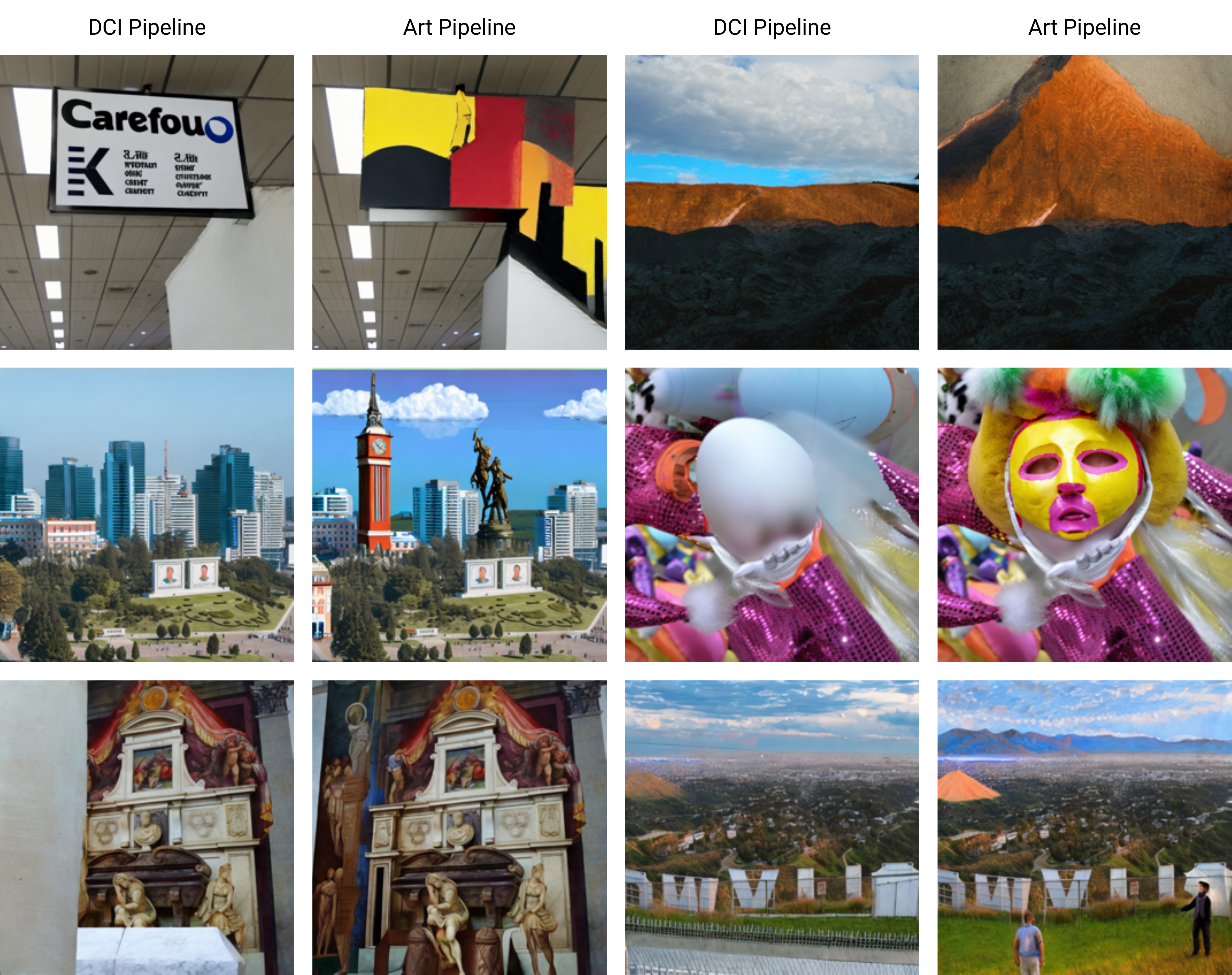}

   \caption{Domain transfer qualitative results.}
   \label{fig:domain_transfer}
\end{figure}

We conducted an additional experiment where we applied the weights learned from the automatically annotated art dataset to the photographic images in the DCI dataset.

As shown in Table \ref{tab:domain_transfer}, while the model trained directly on the DCI dataset achieves higher scores in LPIPS and text-to-image CLIPSim relative to the ground truth annotations, it is slightly outperformed in PSNR and significantly in CLIP-IQA. This latter result suggests that training on the art domain and testing on photographic images can lead to higher-quality outputs in certain metrics.

We analyzed the qualitative results shown in Fig.~\ref{fig:domain_transfer} to better understand these outcomes. These examples illustrate that the model trained on the art dataset tends to apply stylistic features in its completions, often generating inpainted objects reminiscent of paintings from previous centuries, similar to those found in the WikiArt collection. This results in more colorful and aesthetically appealing completions. However, the generated objects are typically less precise, and the overall inpainting shows more artifacts compared to those generated for purely artistic images. This observation indicates that transferring generative capabilities between different domains remains a promising area for further research.

\section{Region-Aligned CLIPSim Computation}

\begin{figure}[t]
  \centering
   \includegraphics[width=1.0\linewidth]{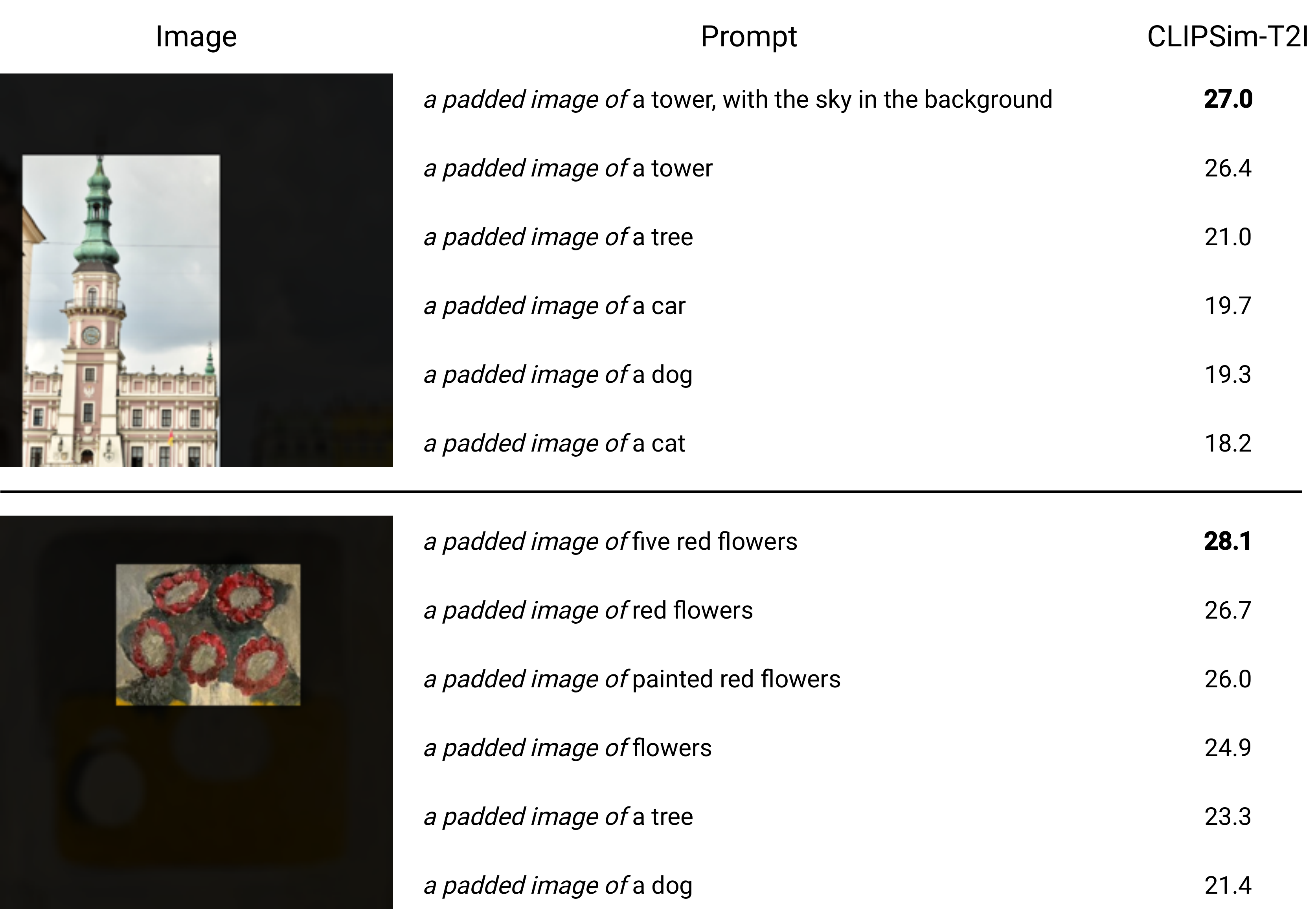}

   \caption{Visualization of our CLIPSim computation to evaluate inpainting prompt-following. By darkening and blurring the rest of the image, we obtain scores aligned to the region of interest.}
   \label{fig:clipsim}
\end{figure}

As detailed in the main paper, we adopt the approach proposed by Lüddecke and Ecker in their work on text-guided segmentation \cite{luddecke2022image} to calculate the CLIP similarity between a regional prompt and the corresponding output. In Fig.~\ref{fig:clipsim}, we visualize how this method is employed as a metric for evaluating the inpainted regions. The results indicate that this metric aligns well with the assessment of prompt adherence in image inpainting tasks.

\section{Additional Qualitative Results}

\begin{figure*}[t]
  \centering
   \includegraphics[width=1.0\textwidth]{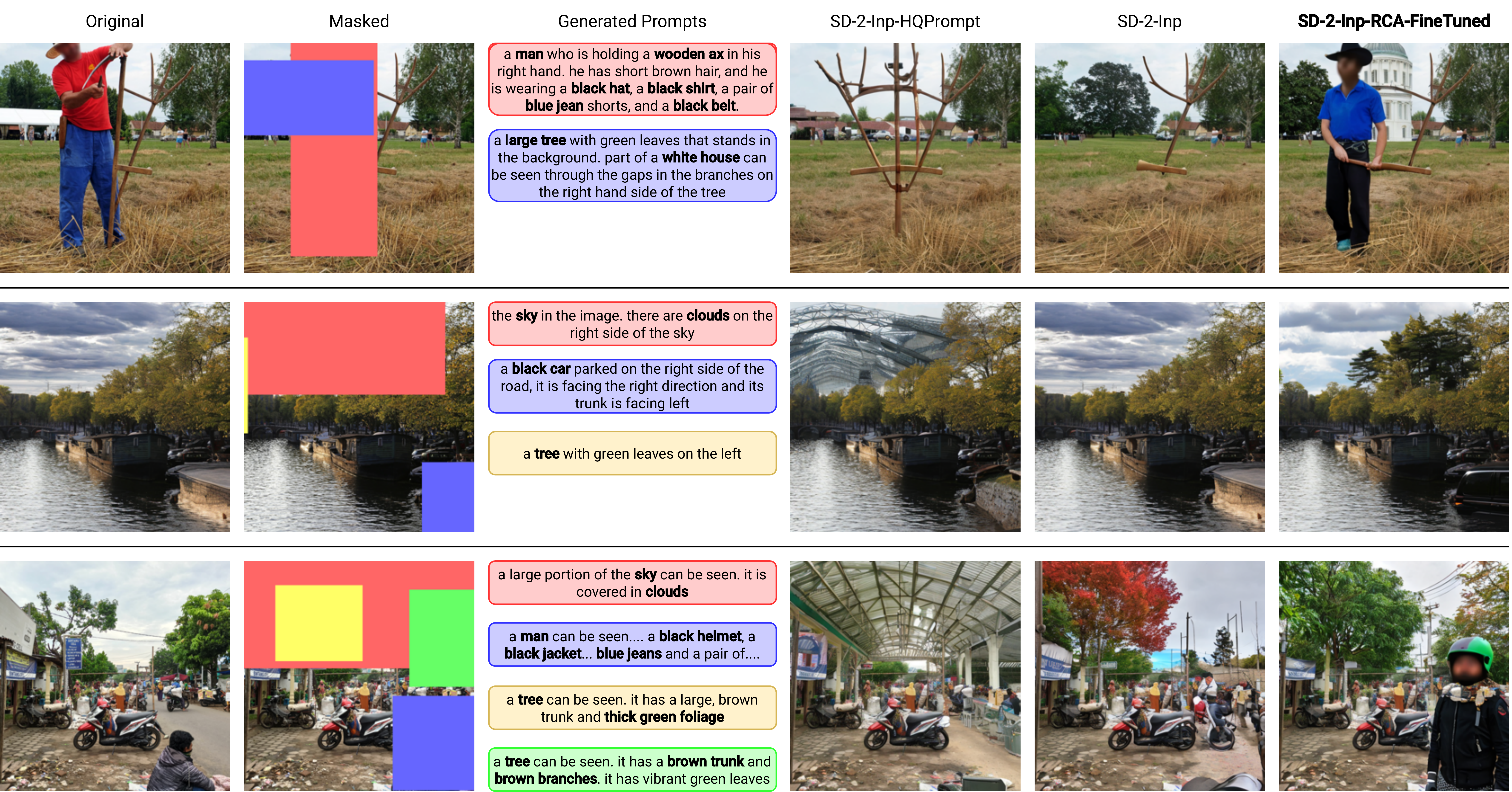}

   \caption{Additional qualitative results on the Densely Captioned Images dataset.}
   \label{fig:qual1}
\end{figure*}

\begin{figure*}[t]
  \centering
   \includegraphics[width=1.0\textwidth]{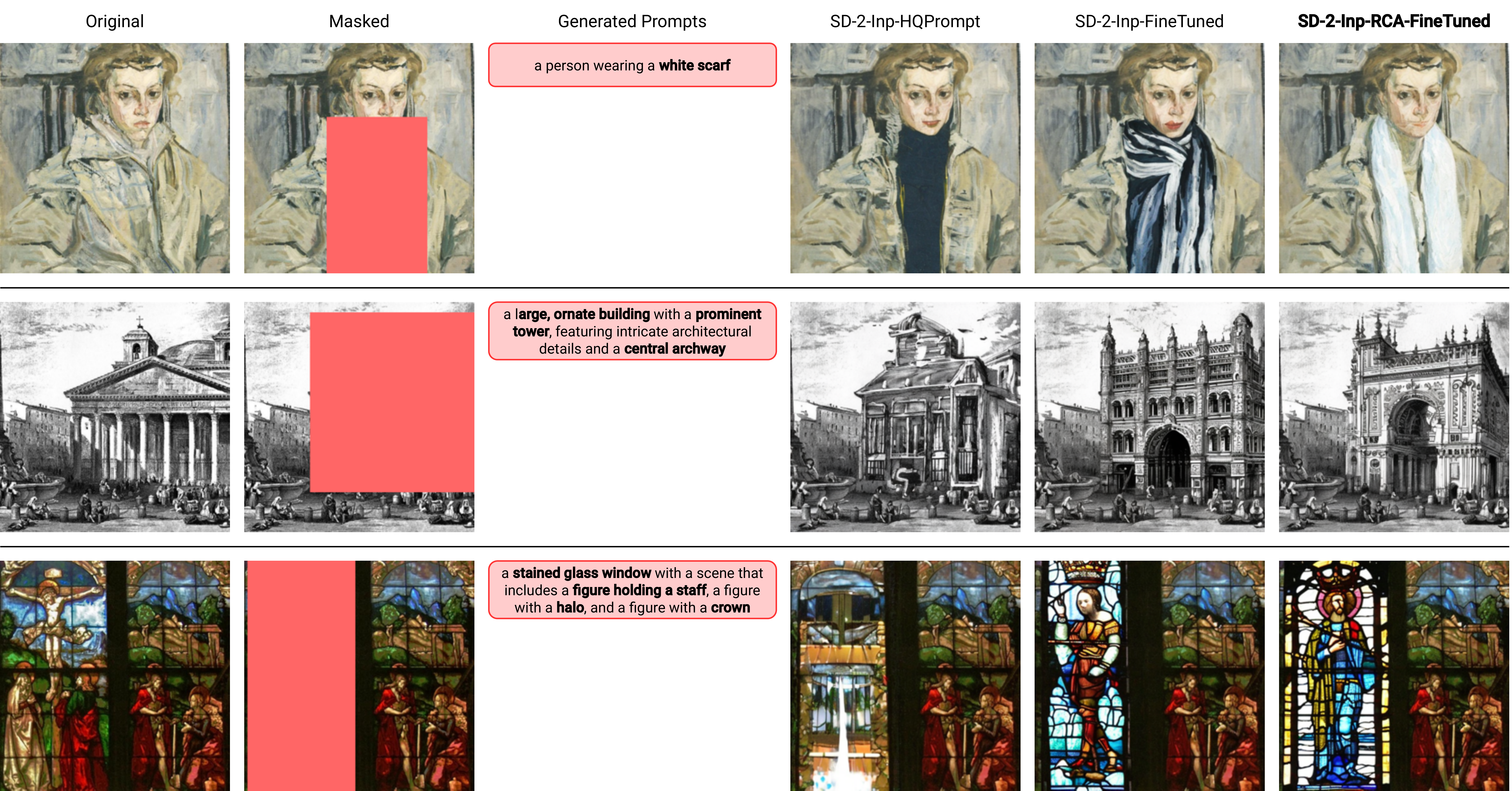}

   \caption{Additional qualitative results on the art dataset for 1-mask inpainting.}
   \label{fig:qual2}
\end{figure*}

\begin{figure*}[t]
  \centering
   \includegraphics[width=1.0\textwidth]{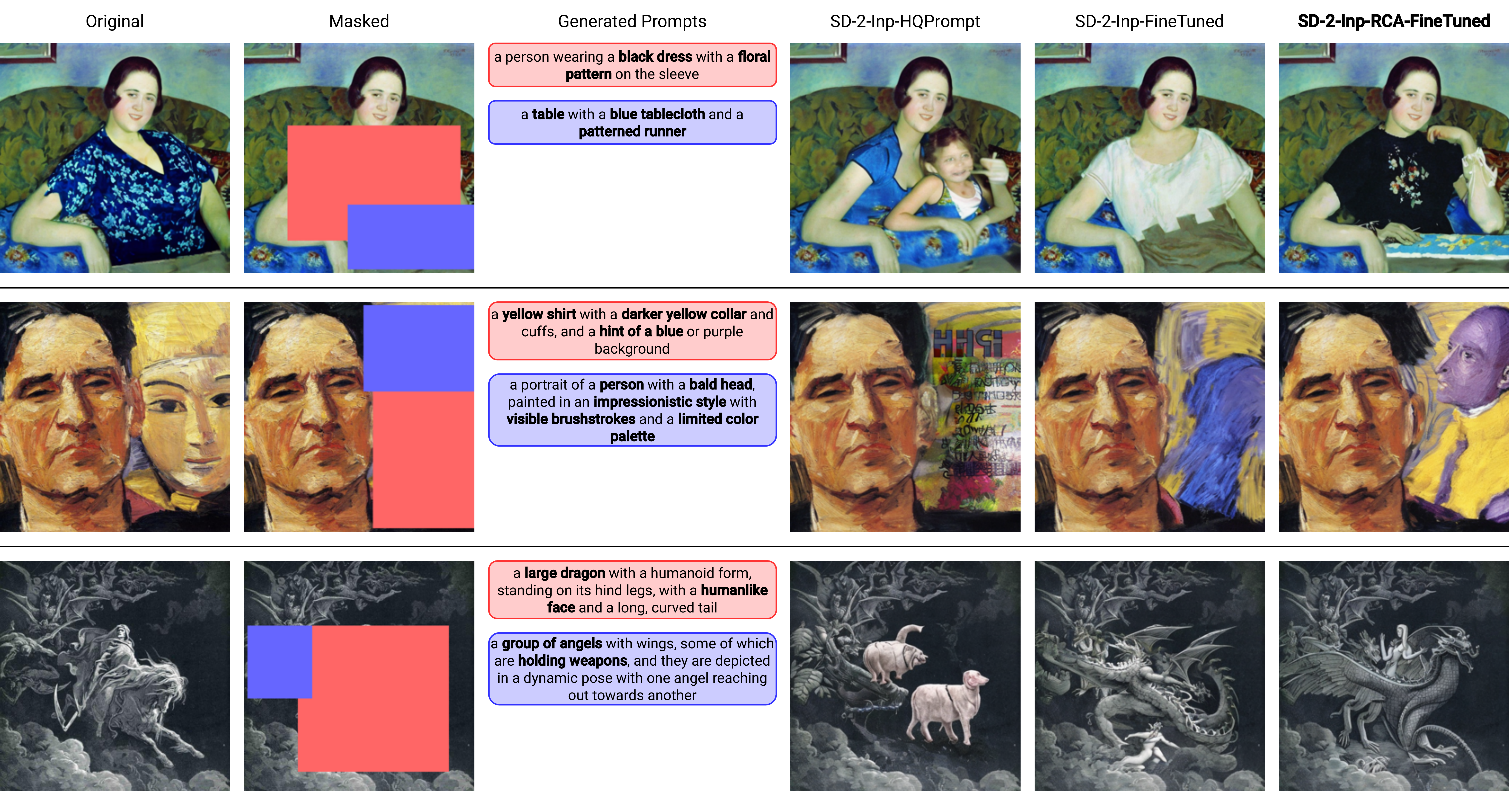}

   \caption{Additional qualitative results on the art dataset for 2-mask inpainting.}
   \label{fig:qual3}
\end{figure*}

\begin{figure*}[t]
  \centering
   \includegraphics[width=1.0\textwidth]{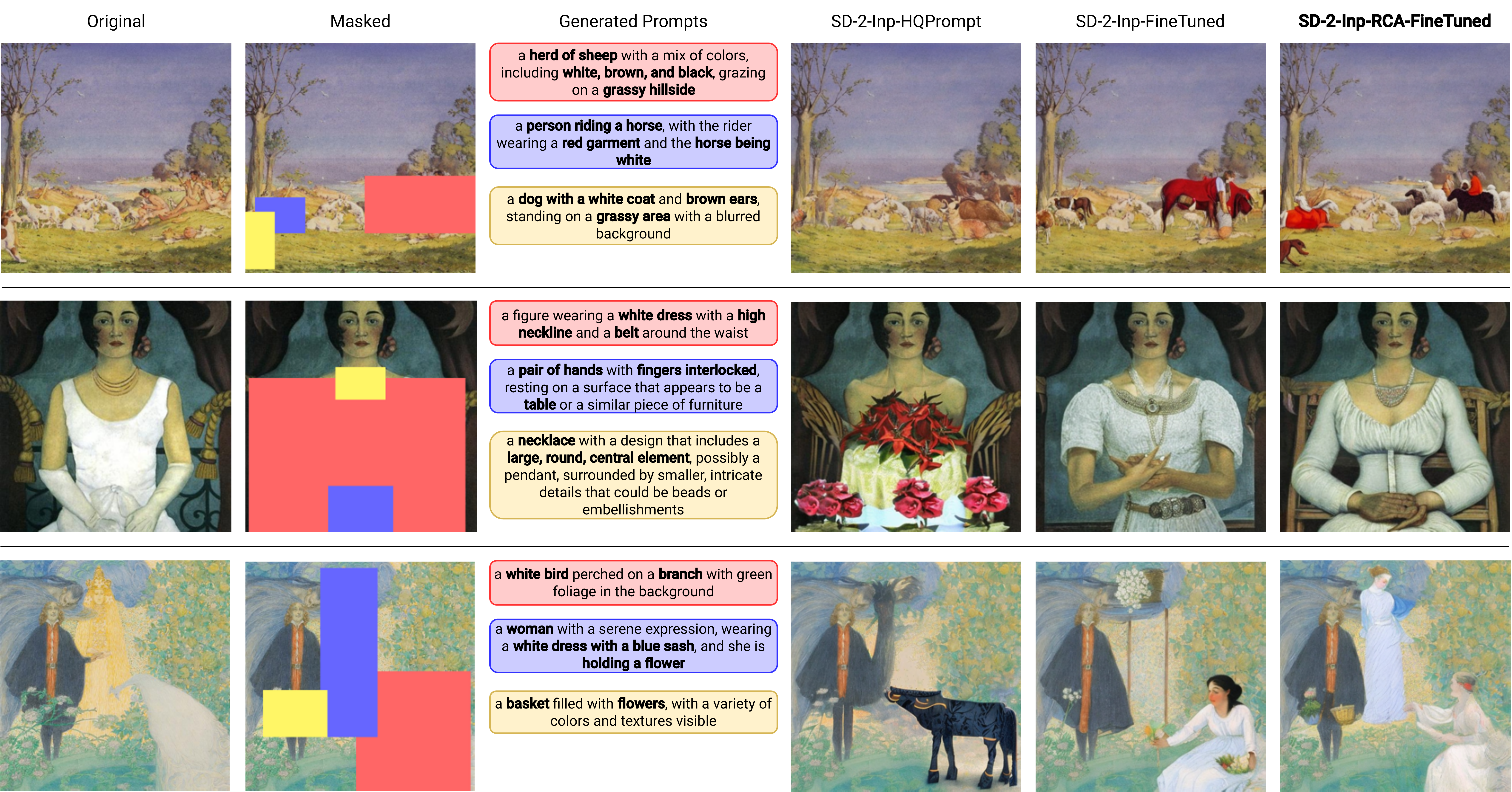}

   \caption{Additional qualitative results on the art dataset for 3-mask inpainting.}
   \label{fig:qual4}
\end{figure*}

\begin{figure*}[t]
  \centering
   \includegraphics[width=1.0\textwidth]{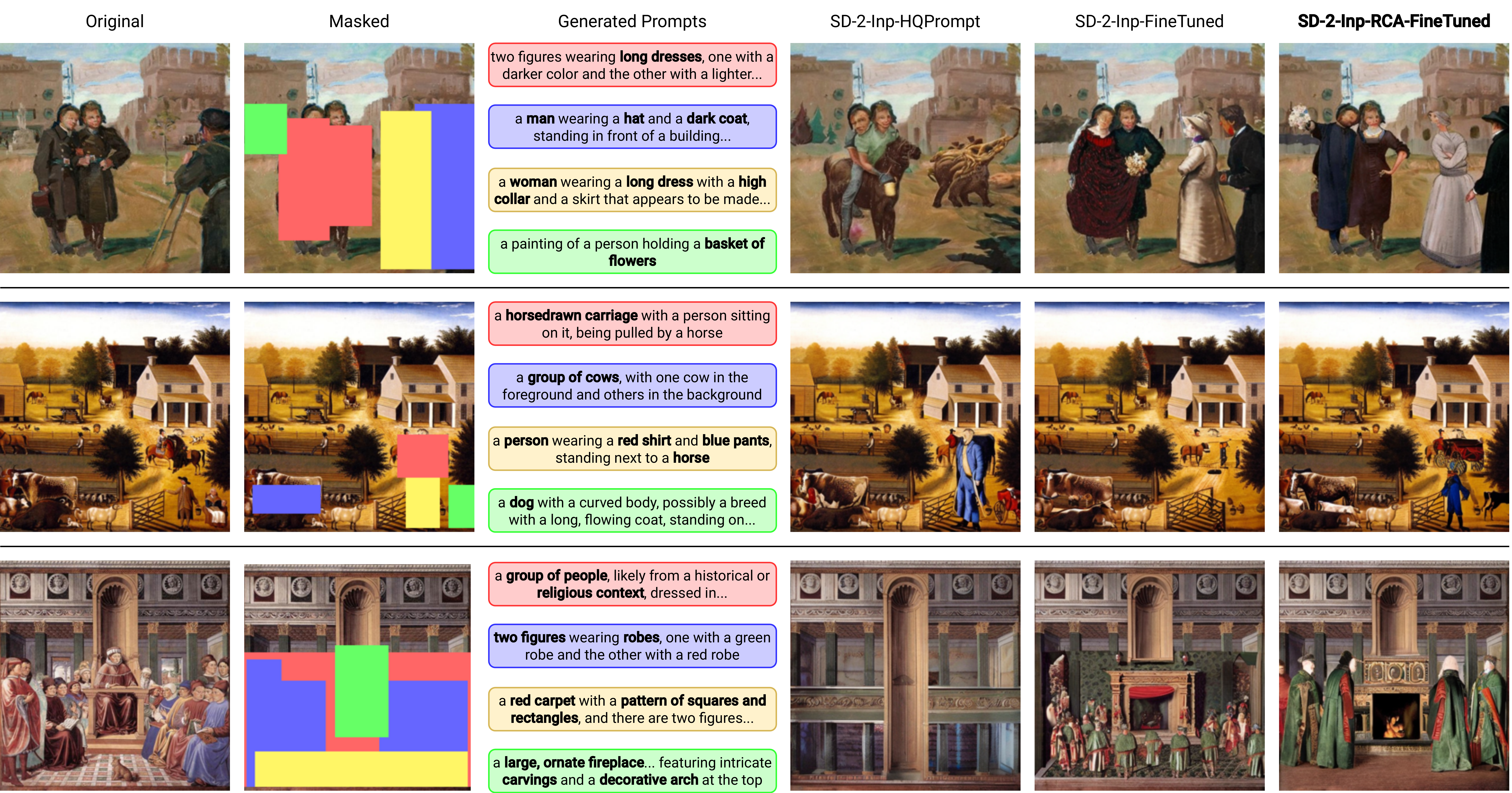}

   \caption{Additional qualitative results on the art dataset for 4-mask inpainting.}
   \label{fig:qual5}
\end{figure*}

\begin{figure*}[t]
  \centering
   \includegraphics[width=1.0\textwidth]{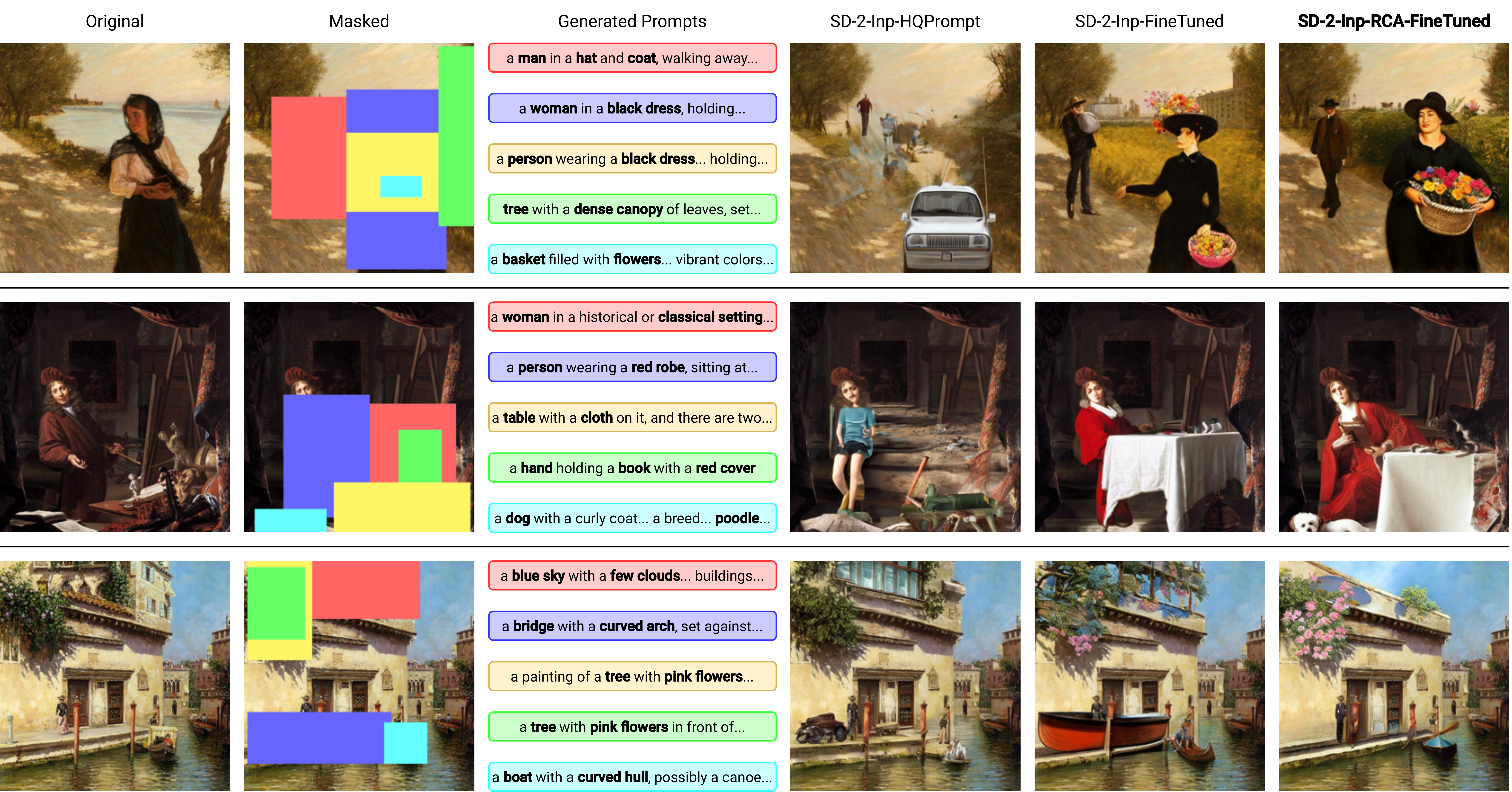}

   \caption{Additional qualitative results on the art dataset for 5-mask inpainting.}
   \label{fig:qual6}
\end{figure*}

We provide additional qualitative results of our pipeline for prompt generation and multi-mask inpainting both on the DCI dataset (Fig.~\ref{fig:qual1}) and on the art dataset, over multiple numbers of inpainting masks (Figs.~\ref{fig:qual2}--\ref{fig:qual6}).

\section{Discussion on Potential Misuse}

In the main paper, we introduced a new pipeline for inpainting multiple regions of an input image using different text prompts. This approach leverages MLLMs and diffusion models, which are currently state-of-the-art in text and image generation. As research advances, these models promise exciting opportunities for creating powerful tools like the one presented here. The results demonstrate that current technology is mature enough to enable MLLMs to interpret images across diverse domains, including complex areas such as artistic images, with high accuracy.

Looking ahead, it is anticipated that MLLMs will increasingly serve as invaluable assistants in various image-related tasks, such as image editing. While some limitations remain—for example, challenges in inpainting large and small areas or managing long prompts, which can sometimes result in noticeable artifacts—ongoing advancements in image generation are expected to continue enhancing these capabilities.

As the technology progresses, the distinction between real and generated images may become increasingly subtle, underscoring the need for innovation and caution. The proposed pipeline provides a practical tool for assisting users, including those with minimal experience in Generative AI, and has potential applications in automating processes such as data augmentation in computer vision. However, remembering the risks, including the accelerated generation of harmful content, is crucial. Therefore, as these tools are refined, it is equally important to develop more effective methods for distinguishing real images from generated ones, thereby helping to mitigate the risk of misuse and ensuring the integrity of digital content.

